%% file: main.tex
\documentclass[compact]{jetbrains_report}

\input{math_commands.tex}

\usepackage{url}
\usepackage{natbib}
\setcitestyle{authoryear,round,citesep={;},aysep={,},yysep={;}}

\usepackage{tabularx}
\usepackage{wrapfig}
\usepackage{xurl}
\usepackage[nameinlink,capitalise,noabbrev]{cleveref}

\title{Budget Internalization: Fixed-Budget RLVR Trades Test-Time Scaling for In-Budget Accuracy}

\jbauthor{Ilia Zavidnyi}{1,2}
\jbauthor{Alexander Kovrigin}{1}
\jbauthor{Arina Puchkova}{1}
\jbauthor{Kirill Gelvan}{1}
\jbauthor{Sergey Titov}{1}
\jbaffiliation{1}{JetBrains Research}
\jbaffiliation{2}{Czech Technical University in Prague}
\jbcorresponding{ilia@zavidnyi.com}

\jbmetadata{Checkpoints}{\href{https://huggingface.co/collections/JetBrains-Research/budget-internalization}{huggingface.co/collections/JetBrains-Research/budget-internalization}}

\newcommand{\fillin}{\textcolor{red}{\textbf{[FILL]}}}
\newcommand{\scoregain}[2]{#1: \textcolor[HTML]{047857}{\(+#2\) pp}}
\newcommand{\scoreloss}[2]{#1: \textcolor[HTML]{B91C1C}{\(-#2\) pp}}
\newcommand{\basegain}[2]{\scoregain{#1}{#2}}
\newcommand{\baseloss}[2]{\scoreloss{#1}{#2}}
\newcommand{\basegainrange}[2]{\textcolor[HTML]{047857}{\(+#1\) to \(+#2\) pp}}
\newcommand{\baselossrange}[2]{\textcolor[HTML]{B91C1C}{\(-#1\) to \(-#2\) pp}}

\begin{document}

\maketitle

\begin{jbabstract}
Reasoning models often become more accurate when they can generate longer answers at test time.
This behavior is known as test-time scaling.
However, RL with verifiable rewards (RLVR) usually trains them with a fixed rollout cap.
We find that this cap changes model behavior, not only training cost.
After training, models use few additional tokens beyond the training budget.
They outperform the base models near that budget, but their accuracy stops improving and falls below the base model at larger budgets.
The rollout cap thus becomes a learned constraint, that limits model's test-time scaling.
We call this effect \emph{budget internalization}.
We observe it across Qwen3.5-4B, Gemma4-E2B, and Nemotron-3-Nano-4B and two reasoning training tasks: mathematical reasoning and diverse, procedurally generated problems.
This effect also transfers from math training to science and coding tasks.
Training-objective level ablations on Qwen3.5-4B show that most changes recover little test-time scaling.
Stronger KL regularization and lower training-data truncation preserve more scaling, but stronger KL reduces in-budget accuracy.
Forcing an answer on truncated training rollouts largely removes budget internalization, but also removes most of the in-budget gain.
SFT on artificially truncated traces reproduces budget internalization, whereas SFT on naturally short traces does not, which shows that the effect is not specific to RL.
This result links budget internalization to length-censored post-training rather than to an RL-specific objective term.
Thus, a rollout cap does not only control training cost.
It can trade low-budget gains for lower performance when additional inference compute is available.
\end{jbabstract}

\begin{figure}[h]
    \centering
    \vspace{8pt} 
    \includegraphics[width=\textwidth]{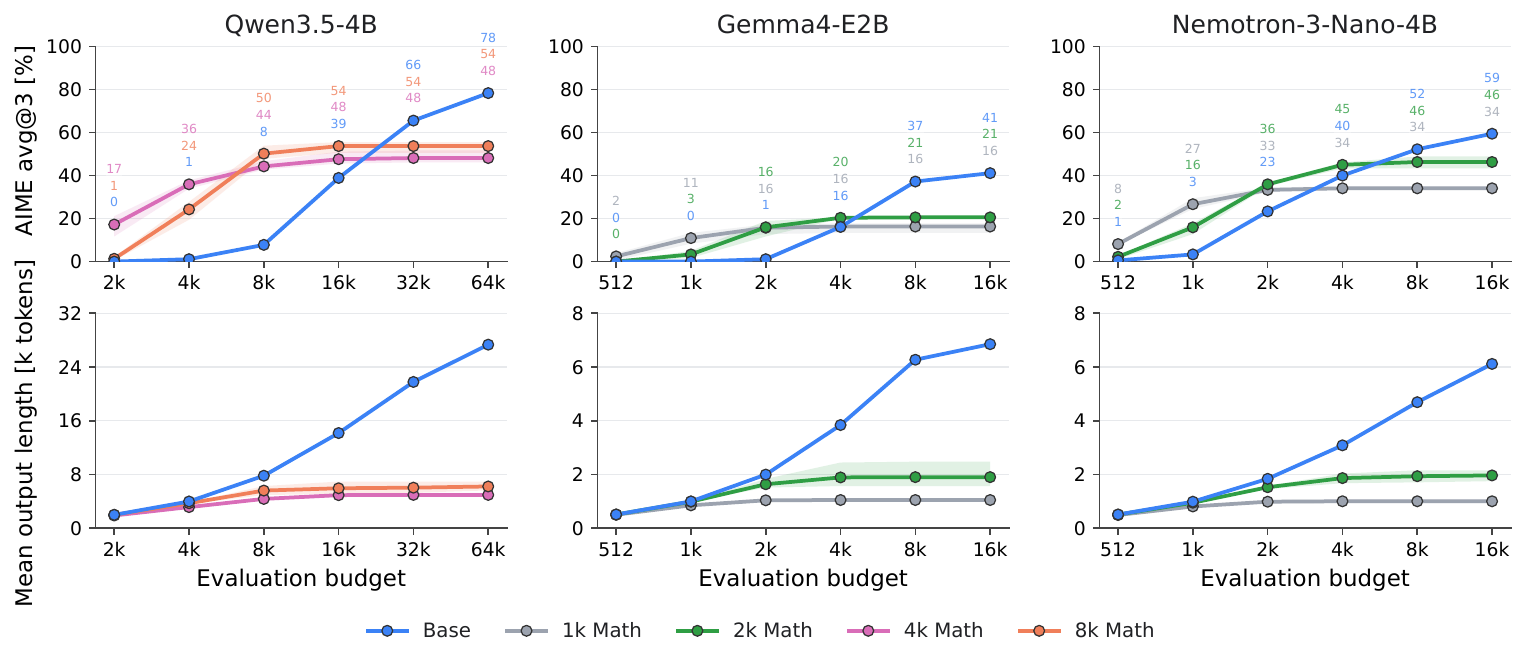}
    \caption{\emph{Budget internalization} after fixed-budget math RLVR.
    Trained points and labels are means over three validation-selected checkpoints; shading shows the seed minimum--maximum.
    Blue curves are base-model evaluations. Appendix~\ref{app:multiseed} reports selected and final results for each seed.}
    \label{fig:headline}
\end{figure}

\clearpage 
\input{introduction}
\input{method}
\input{analysis}
\input{literature_review}
\input{conclusion}
\newpage

\subsection*{AI use statement}

In this work, we used generative AI tools to assist with language editing, code development, figure generation, and document formatting.
All research decisions, experiments, numerical results, citations, and AI-assisted outputs were reviewed and verified by the authors.
The authors take responsibility for the final content of the paper.

\subsection*{Reproducibility statement}

We release all trained checkpoints, including ablations, at \url{https://huggingface.co/collections/JetBrains-Research/budget-internalization}.
Appendix~\ref{app:setup} reports the complete training and evaluation configuration.
The remaining appendices provide the evaluation prompts, exact values underlying the main figures, and additional results needed to reproduce the reported comparisons.

\bibliography{references}
\bibliographystyle{references}

\newpage
\appendix
\suppressfloats[t] 
\crefalias{section}{appendix}
\crefalias{subsection}{appendix}
\input{appendix}
\clearpage
\input{multiseed}

\end{document}

%% file: math_commands.tex
\usepackage{amsmath,amsfonts,bm}

\def\eqref#1{equation~\ref{#1}}

\def\1{\bm{1}}

\DeclareMathAlphabet{\mathsfit}{\encodingdefault}{\sfdefault}{m}{sl}
\SetMathAlphabet{\mathsfit}{bold}{\encodingdefault}{\sfdefault}{bx}{n}



%% file: introduction.tex
\section{Introduction}

\begingroup
\renewcommand\thefootnote{}
\footnotetext{We release all our trained checkpoints (including ablations) at \url{https://huggingface.co/collections/JetBrains-Research/budget-internalization}.}
\endgroup

One promise of reasoning models is that they can solve harder problems when given more tokens at test time \citep{deepseek2025r1, snell2024scaling}.
This makes test-time compute a useful scaling axis: a deployment can spend more compute when a problem is difficult or when higher accuracy is required.

Unlike test-time scaling, RL with verifiable rewards (RLVR) usually does not let the model reason until it emits an end-of-sequence token.
Instead, training uses a fixed rollout cap, often below the length of the model could otherwise utilize\citep{kimi2025k15, luo2025deepscaler, he2025skywork}.
This is usually done to cut training costs, since it can scale quadratically with the sequence length \citep{aghajohari_markovian_2025}.
The resulting model is then often evaluated at only one budget, at or close to the training cap.
This evaluation shows whether training helps near that budget, but not whether the model can still benefit from more compute.
A model can look better than its base model at the training budget and become worse when both receive a larger budget.

We show that this failure occurs after fixed-budget GRPO on two hard reasoning training tasks: competition mathematics (AIME, DeepScaler~\cite{luo2025deepscaler}) and diverse procedurally generated reasoning problems (Reasoning Gym~\cite{reasoninggym2025}).
For Qwen3.5-4B trained with a 4k rollout cap, the model moves from a large gain over the base model at the matching evaluation budget to a large deficit when given more tokens (\basegain{4K}{34.8}; \baseloss{64K}{30.2}, averaged over three training seeds).
Gemma4-E2B~\citep{gemmateam2026gemma4} and Nemotron-3-Nano-4B~\citep{nvidia2026nemotron3nano} show the same saturation after math training, and Reasoning Gym training reproduces it for all three models.
A similar pattern also transfers from math training to GPQA-Diamond \citep{rein2023gpqagraduatelevelgoogleproofqa} and LiveCodeBench \citep{jain2024livecodebench}.
We call this effect \emph{budget internalization}: the training budget becomes part of the model's learned behavior rather than remaining only a limit on training cost.

Budget internalization changes how RLVR results should be interpreted.
Accuracy measured only at the training cap can hide a reversal in model ranking.
Increasing the deployment budget may then fail to recover the base model's performance, even though the base model benefits from the same additional compute.
The training cap can therefore trade future test-time scaling for better performance at the budget used during training.

Our goal is to investigate budget internalization and understand why it occurs.
To do this, we measure accuracy and output-length curves across model families, training caps, and reasoning benchmarks.
We then run training ablations to study which training choices affect above-budget scaling.
This analysis makes three contributions:

\begin{itemize}
\item \textbf{We show that fixed-budget training hurts test-time scaling.}
Across Qwen, Gemma, and Nemotron, AIME accuracy improves near the training cap but stops improving at larger budgets.
Across all six checkpoints, gains over the base model range from \basegainrange{11.7}{46.7} at the training cap, while deficits at the largest evaluation budget shown for each model range from \baselossrange{17.2}{35.0}.
\item \textbf{The effects of fixed-budget RLVR transfer beyond the training task and can improve efficiency.}
For Qwen's 8k-trained checkpoint, the high-budget differences from the base model are much smaller (\baseloss{GPQA, 64K}{2.2}; \baseloss{LCB-v6, 64K}{1.0}).
At the same time, mean output length falls compared to the base Qwen3.5-4B from \(11.15\)k to \(4.56\)k tokens on GPQA and from \(38.32\)k to \(8.31\)k on LiveCodeBench.
Gemma and Nemotron show the same pattern of much shorter responses with smaller accuracy differences on these tasks.
Therefore, losing some test-time scaling is not always harmful and can reduce inference cost while retaining similar accuracy.
\item \textbf{Our ablations reveal a persistent trade-off between in-budget gains and test-time scaling.}
We examine objective variants, truncation handling, KL regularization, and SFT approaches.
Scaling remains impaired under alternative length normalization and overlength penalties, but improves when truncation exposure is reduced or KL anchoring is strengthened.
The scoring of truncated rollouts strongly affects budget internalization, suggesting that the responses reinforced during training matter more than any single GRPO objective term.
SFT experiments support this interpretation: scaling is retained with naturally completed trajectories, but deteriorates with artificially shortened ones.
Despite these partial mitigations, no tested intervention combines the full in-budget improvement with the base model's high-budget performance.
\end{itemize}

Budget internalization is therefore neither always good nor always bad.
It is harmful when a task requires longer reasoning to solve harder problems, but useful when the same accuracy can be reached with fewer tokens on tasks where the model tends to overthink.
Evaluating only at the training cap cannot distinguish these cases.
We therefore recommend reporting accuracy and output length at several budgets, so that the effect of fixed-budget training is clear before choosing a model for deployment.

%% file: method.tex
\section{How Fixed-Budget Training Changes Test-Time Scaling}
\label{sec:main}

\label{sec:setup}

We study Qwen3.5-4B, Gemma4-E2B, and NVIDIA-Nemotron-3-Nano-4B.
We run GRPO on the DeepScaleR math dataset with 4k and 8k rollouts cap for Qwen, and 1k and 2k caps for Gemma and Nemotron.
We additionally train Qwen with 4k cap and Gemma and Nemotron with 2k caps on \href{https://huggingface.co/datasets/nvidia/Nemotron-RL-ReasoningGym-v1}{Nemotron-RL-ReasoningGym-v1}~\citep{nvidia2026reasoninggym}, which contains 15,000 examples from 104 procedurally generated and verifiable Reasoning Gym environments~\citep{reasoninggym2025}.
These runs use the same training hyperparameters as the corresponding DeepScaleR runs.
We choose these caps below the budgets at which the base models usually complete their AIME responses, so base-model generations are frequently truncated at every chosen cap; Appendix Table~\ref{tab:truncation-rates} reports the exact rates.
We evaluate the selected checkpoints on AIME 2024 and 2025.
We then test whether the resulting behavior transfers to out-of-domain benchmarks, GPQA-Diamond and LiveCodeBench v6.
All runs use the NeMo RL framework \citep{nemo-rl}; the reward verifies the final answer and assigns zero reward to rollouts that end at the cap without one.

We run all evaluations with the Inspect AI framework \citep{uk_ai_security_institute_2024_inspect_ai} and report avg@k, the accuracy averaged over $k$ sampled generations per problem, with the evaluation budget as the generated-token limit.
For DeepScaleR/AIME, each model/cap condition uses seeds 42, 43, and 44, evaluated at both its in-budget validation-selected checkpoint and final step 300 (Appendix~\ref{app:multiseed}).
Other experiments retain the original single-seed checkpoints and base-model evaluations.
Appendix~\ref{app:setup} reports the complete training and evaluation configuration.

\subsection{Fixed-Budget Training Limits Test-Time Scaling}

\begin{wraptable}{r}{0.40\textwidth}
\centering
\footnotesize
\setlength{\tabcolsep}{2.5pt}
\caption{Mean AIME output length at each model's largest budget in Figure~\ref{fig:headline} (Qwen: 64k; Gemma and Nemotron: 16k). Trained entries are seed mean [min, max]; bases use one evaluation.}
\label{tab:above-cap-saturation}
\begin{tabular*}{\linewidth}{@{\extracolsep{\fill}}lr@{}}
\toprule
Checkpoint & Mean length \\
\midrule
\textbf{Qwen3.5-4B} base & 27.32 \\
\quad 4k-trained & 4.97 [{\scriptsize 4.71, 5.25}] \\
\quad 8k-trained & 6.22 [{\scriptsize 5.64, 6.95}] \\
\addlinespace[2pt]
\textbf{Gemma4-E2B} base & 6.85 \\
\quad 1k-trained & 1.05 [{\scriptsize 0.99, 1.11}] \\
\quad 2k-trained & 1.90 [{\scriptsize 1.57, 2.48}] \\
\addlinespace[2pt]
\textbf{Nemotron-3-Nano-4B} base & 6.11 \\
\quad 1k-trained & 1.01 [{\scriptsize 0.93, 1.08}] \\
\quad 2k-trained & 1.97 [{\scriptsize 1.75, 2.16}] \\
\bottomrule
\end{tabular*}
\end{wraptable}

Figure~\ref{fig:headline} reports seed means and minimum--maximum bands.
All 18 runs reproduce the crossover under both checkpoint choices (Appendix~\ref{app:multiseed}).
Accuracy deltas below compare seed means with the base model at the same budget.
Training gives large gains where the base model frequently reaches the evaluation cap, but these gains quickly saturate once the evaluation budget exceeds the training budget.
Both Qwen checkpoints show this reversal: the 4k-trained checkpoint moves from \basegain{4K}{34.8} to \baseloss{64K}{30.2}, while the 8k-trained checkpoint moves from \basegain{8K}{42.4} to \baseloss{64K}{24.6}.
The same pattern holds for both Gemma checkpoints: the 1k-trained checkpoint moves from \basegain{1K}{10.9} to \baseloss{16K}{24.8}, while the 2k-trained checkpoint moves from \basegain{2K}{14.8} to \baseloss{16K}{20.6}.
For Nemotron, the 1k-trained checkpoint moves from \basegain{1K}{23.3} to \baseloss{16K}{25.4}, while the 2k-trained checkpoint moves from \basegain{2K}{12.6} to \baseloss{16K}{13.1}.

The output-length curves show that this is not only an accuracy plateau.
The trained models rarely use the full additional token allowance.
Table~\ref{tab:above-cap-saturation} reports the mean and sed range of mean output length at these endpoints.
The base models average 6.11k--27.32k tokens, whereas the trained checkpoints average only 1.01k--6.22k tokens.
Smaller training caps therefore shorten the model's overall response distribution rather than imposing a hard decoding limit.

\subsection{Budget Internalization Across Training Tasks}

\begin{figure}[h!]
\centering
\includegraphics[width=\textwidth]{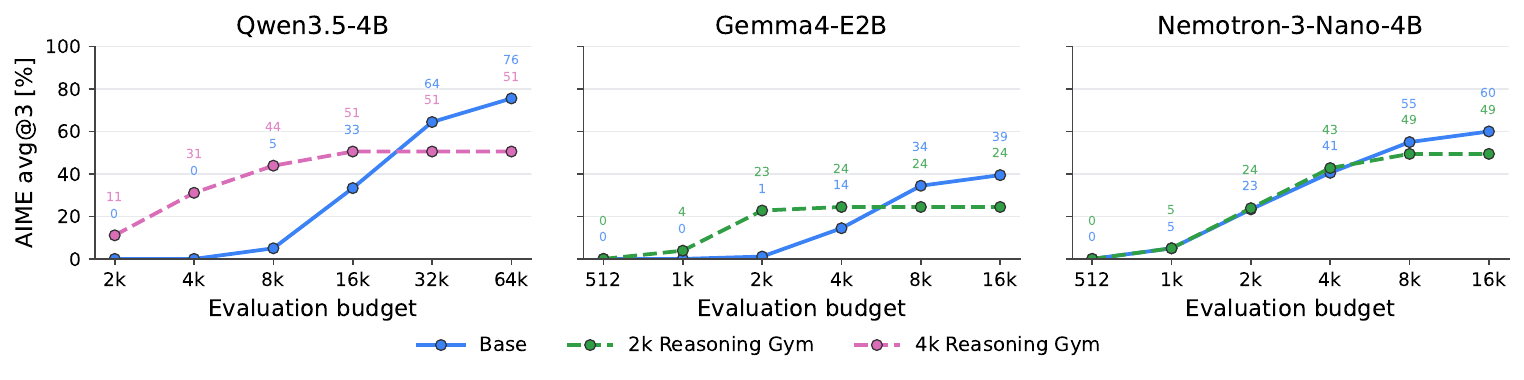}
\caption{AIME accuracy after fixed-budget Reasoning Gym training across three model families, with dashed lines for trained checkpoints and colors matching Figure~\ref{fig:headline}.
Appendix Table~\ref{tab:reasoning-gym-exact-values} reports the exact values.}
\label{fig:reasoning-gym-training}
\end{figure}

We next test whether this pattern is specific to mathematical-reasoning training by replacing DeepScaleR with the diverse procedural tasks in \href{https://huggingface.co/datasets/nvidia/Nemotron-RL-ReasoningGym-v1}{Nemotron-RL-ReasoningGym-v1}~\citep{nvidia2026reasoninggym}.
Figure~\ref{fig:reasoning-gym-training} shows that it is not.
For Qwen, the 4k-trained checkpoint moves from \basegain{4K}{31.1} to \baseloss{64K}{25.0}.
Gemma's 2k-trained checkpoint moves from \basegain{2K}{21.7} to \baseloss{16K}{15.0}.
Nemotron shows a smaller in-budget effect: its 2k-trained checkpoint gains only \basegain{2K}{0.6}, but it then saturates at \(49.4\%\) while the base model reaches \(60.0\%\) at 16k (\baseloss{16K}{10.6}).
The in-budget gain therefore depends on the model and training task, but the loss of test-time scaling appears after fixed-budget training on both hard reasoning datasets.

\subsection{Transfer to Other Tasks and Budget Forcing}

\begin{figure}[h!]
\centering
\includegraphics[width=\textwidth]{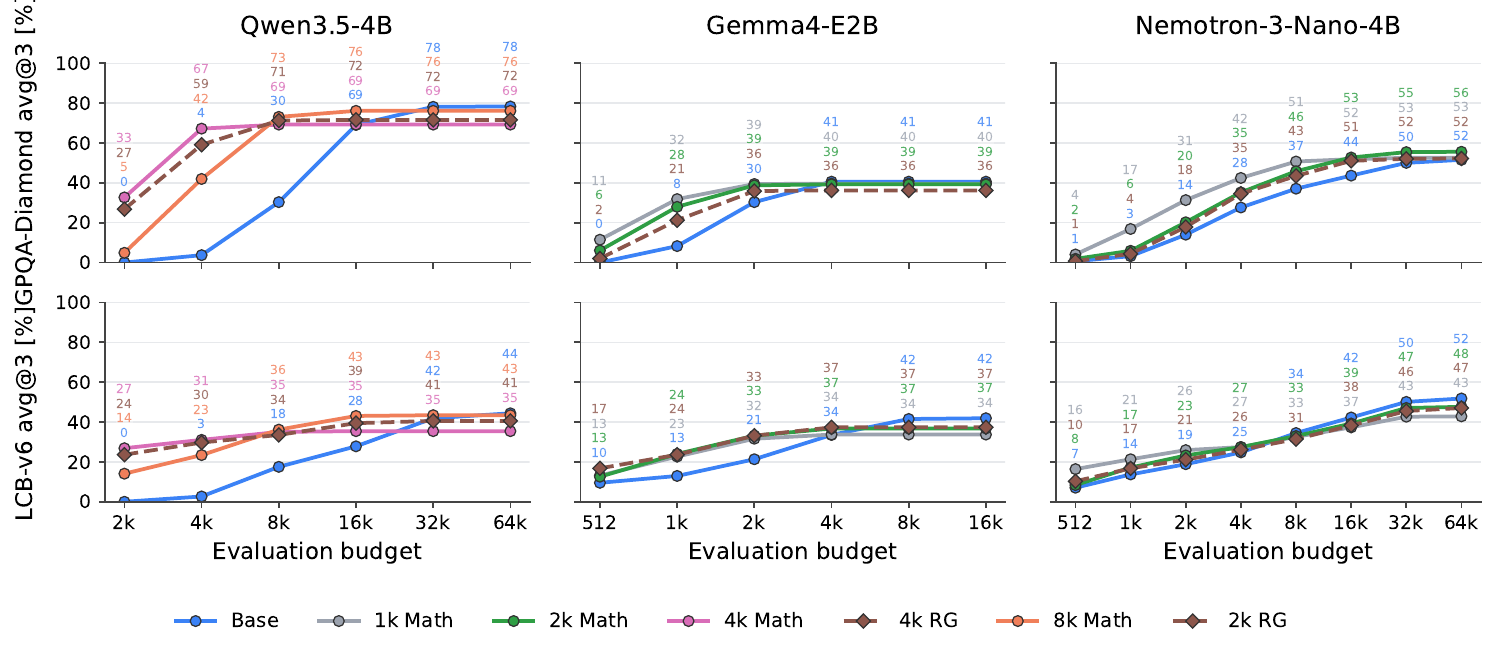}
\caption{Accuracy on GPQA-Diamond and LiveCodeBench v6 after fixed-budget math (solid budget colors) and Reasoning Gym (brown dashed diamonds) training.
Appendix Tables~\ref{tab:transfer-exact-values} and~\ref{tab:reasoning-gym-transfer-exact-values} report exact values; Appendix Figure~\ref{fig:transfer-lengths} reports DeepScaleR output lengths.}
\label{fig:transfer}
\end{figure}

The behavior transfers from both training datasets to GPQA-Diamond, a graduate-level science benchmark, and LiveCodeBench v6, a code-generation benchmark with hidden tests (\Cref{fig:transfer}).
The trained checkpoints gain accuracy near their training budget and saturate while producing shorter responses, although the high-budget damage is smaller than on AIME.
On GPQA and LiveCodeBench, both Qwen checkpoints preserve much more of the base model's high-budget accuracy than on AIME (GPQA at 64k: \baseloss{4k-trained}{9.1}, \baseloss{8k-trained}{2.2}; LCB-v6 at 64k: \baseloss{4k-trained}{9.0}, \baseloss{8k-trained}{1.0}).
Gemma's 1k-trained checkpoint follows the same pattern on GPQA (\baseloss{16K}{1.0}) and LiveCodeBench (\baseloss{16K}{8.2}).
Its 2k-trained checkpoint also remains close to the base model on GPQA (\baseloss{16K}{1.3}) and reduces the LiveCodeBench deficit (\baseloss{16K}{5.1}).
Nemotron's 1k-trained checkpoint exceeds the base model on GPQA at high budget (\basegain{64K}{1.0}) but not on LiveCodeBench (\baseloss{64K}{9.0}).
Its 2k-trained checkpoint strengthens the GPQA improvement (\basegain{64K}{4.0}) and narrows the LiveCodeBench deficit (\baseloss{64K}{4.2}).
Reasoning Gym checkpoints follow the same pattern: Qwen remains below the base model on GPQA (\baseloss{64K}{6.7}) and LiveCodeBench (\baseloss{64K}{3.8}), as does Gemma (\baseloss{GPQA, 16K}{4.4}; \baseloss{LCB-v6, 16K}{4.6}), while Nemotron slightly exceeds the base model on GPQA (\basegain{64K}{0.7}) but not on LiveCodeBench (\baseloss{64K}{4.8}).
Because these differences are smaller, Appendix~\ref{app:pass-at-three} additionally reports pass@3 curves for a more stable comparison.
We hypothesize that these tasks often require less extended reasoning, so the base model sometimes overreasons rather than making productive use of additional tokens. 

\label{budget-forcing-desc}
This behavior makes the low-budget comparison difficult to interpret.
At low budgets, the base model often overreasons until it reaches the cap and fails to produce a final answer.
Table~\ref{tab:base-truncation-rates} shows that this is common across all three benchmarks.

\begin{table}[h!]
\centering
\footnotesize
\setlength{\tabcolsep}{5pt}
\caption{Qwen3.5-4B base-model truncation rate (\%) across evaluation budgets.
Appendix~\ref{app:truncation-rates} reports the corresponding curves and exact values for every model family and fixed-budget checkpoint.}
\label{tab:base-truncation-rates}
\begin{tabular*}{\textwidth}{@{\extracolsep{\fill}}lrrrrrr@{}}
\toprule
Task & 2k & 4k & 8k & 16k & 32k & 64k \\
\midrule
AIME & 100.00 & 100.00 & 95.00 & 66.67 & 33.33 & 6.67 \\
GPQA-Diamond & 100.00 & 95.79 & 67.00 & 16.67 & 0.51 & 0.00 \\
LiveCodeBench v6 & 99.81 & 97.14 & 81.52 & 70.86 & 51.81 & 45.14 \\
\bottomrule
\end{tabular*}
\end{table}
We therefore first test whether the base model could answer correctly without continuing to overreason.
We apply the budget-forcing procedure from \citet{muennighoff2025s1}.
If a response reaches budget \(B\) without a parseable answer, we retain its reasoning prefix, insert the model-specific end-of-thinking delimiter and answer cue, and allow a short answer window.
Naturally completed responses remain unchanged.
Examples showing the prompt, model response, and budget-forcing intervention for all models and benchmarks are provided in \Cref{app:budget-forcing-prompts}.

\begin{figure}[h!]
\centering
\includegraphics[width=\textwidth]{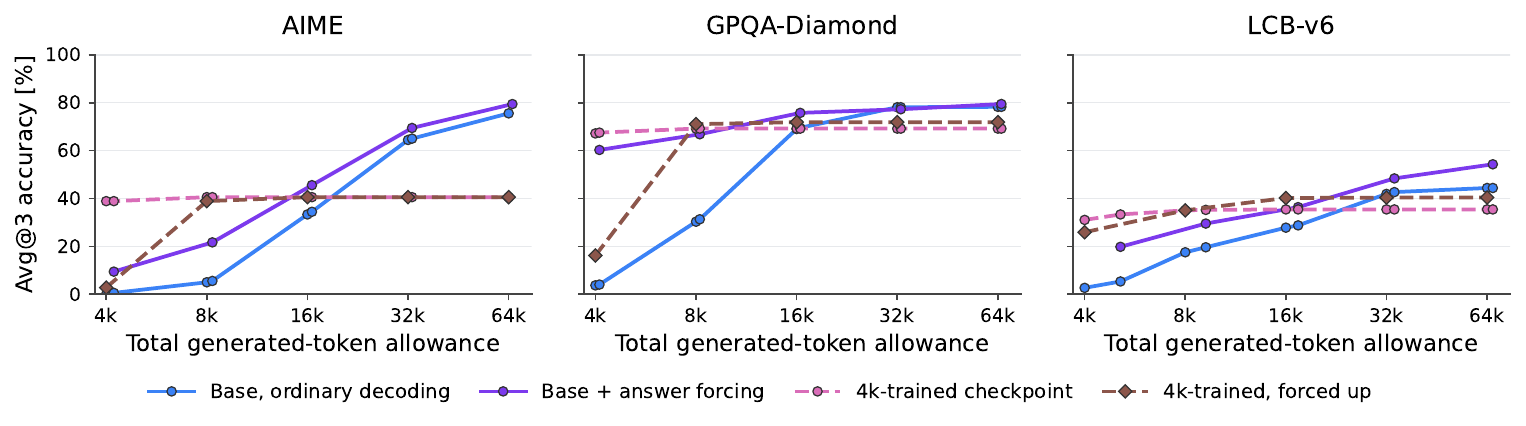}
\caption{Qwen3.5-4B accuracy under ordinary base-model decoding, base-model answer forcing, the 4k-trained checkpoint, and forcing the trained checkpoint upward.
Appendix Figure~\ref{fig:forcing-diagnostics} and Table~\ref{tab:forcing-diagnostics} report truncation rates and output lengths.}
\label{fig:budget-forcing-qwen}
\end{figure}

Budget forcing substantially improves low-budget accuracy across all three tasks (\basegain{AIME, 4K}{8.9}; \basegain{GPQA, 4K}{56.2}; \basegain{LCB-v6, 4K}{14.5}; \Cref{fig:budget-forcing-qwen}).
The same trend persists at larger budgets, where answer forcing generally preserves or improves accuracy.
Appendix Figures~\ref{fig:budget-forcing-gemma} and~\ref{fig:budget-forcing-nemotron} report the corresponding base-model intervention for Gemma and Nemotron.

Next, we apply the reverse intervention to the 4k-trained checkpoint to test whether its loss of test-time scaling only reflects early stopping.
We remove its end-of-thinking delimiter and final answer, append \texttt{Wait}, and ask it to continue the same trace, repeating this intervention up to four times under a 64k cap.
Relative to ordinary decoding of the same checkpoint, forcing up lowers accuracy at the 4k training budget (\scoreloss{AIME, 4K}{36.1}; \scoreloss{GPQA, 4K}{51.0}; \scoreloss{LCB-v6, 4K}{5.1}) and produces the same or only modestly higher accuracy at 64k (AIME, 64K: \(0.0\) pp; \scoregain{GPQA, 64K}{2.7}; \scoregain{LCB-v6, 64K}{5.0}), despite using \(2.58\)--\(2.74\)k more tokens (\Cref{app:add-wait}).
Stating the budget explicitly in the prompt offers no reliable alternative, as models not trained on such cues largely ignore them (\Cref{sec:conditioning}).

Budget forcing can therefore improve base-model accuracy at a matched total token allowance, while making the trained model spend more tokens does not restore its test-time scaling.
This suggests that fixed-budget training genuinely compresses the reasoning needed to reach its answers rather than only teaching the model to stop early.
Appendix~\ref{app:budget-forcing-prompts} gives complete examples of both procedures, while Appendix Figures~\ref{fig:budget-forcing-gemma} and~\ref{fig:budget-forcing-nemotron} report the corresponding results showing the same pattern for Gemma and Nemotron.

\subsection{Budget Internalization}
Fixed-budget training therefore has three distinct effects: it can improve low-budget accuracy, make reasoning more economical, and reduce how much the model benefits from additional inference compute.
We call the last effect \emph{budget internalization}.
A model internalizes its training budget when, above that budget, its base model continues to use more tokens and improve with additional compute while the trained model's response length and accuracy largely saturate.

%% file: analysis.tex
\section{What Affects Budget Internalization?}
\label{sec:mechanism}

\newcommand{\hypothesisquestion}[1]{\textbf{#1}}
\newcommand{\motivationlabel}{\textcolor[HTML]{047857}{\textbf{Motivation.}}}
\newcommand{\testlabel}{\textcolor[HTML]{B45309}{\textbf{Test.}}}
\newcommand{\resultslabel}{\textcolor[HTML]{7E22CE}{\textbf{Results.}}}
\newcommand{\answerlabel}{\textcolor[HTML]{B91C1C}{\textbf{Answer.}}}
\newcommand{\bettervalue}[1]{\textcolor[HTML]{047857}{#1}}
\newcommand{\worsevalue}[1]{\textcolor[HTML]{B91C1C}{#1}}
\newcommand{\bestvalue}[1]{\textbf{#1}}
\newcommand{\secondvalue}[1]{\underline{#1}}
\newcommand{\baserow}{\rowcolor[gray]{0.92}}

\newlength{\ciwidth}
\settowidth{\ciwidth}{\tiny\(\pm00.0\)}
\newcommand{\civalue}[1]{\makebox[\ciwidth][l]{\tiny\textcolor[gray]{0.45}{\(\pm#1\)}}}

Having established this behavior, we now inspect the potential causes of budget internalization.
We hypothesize that it arises from how the training objective treats truncated rollouts.
To understand where this behavior may come from, let us inspect the GRPO objective and identify where the generation budget enters training \citep{shao2024deepseekmath}:
\begin{equation}
\mathcal{J}(\theta)
=
\mathbb{E}_{q \sim \mathcal{D}}
\mathbb{E}_{\{o_i\} \sim \pi_{\theta_{\mathrm{old}}}(\cdot \mid q;B,\Pi)}
\left[
w_B(q,\{o_i\})\sum_i c_B(|o_i|)\ell_i
-\beta D_{\mathrm{KL}}(\pi_\theta\Vert\pi_{\mathrm{ref}})
\right].
\label{eq:decomp}
\end{equation}
The budget \(B\) first enters the rollout policy by limiting how long a response can be generated, while the termination protocol \(\Pi\) determines what happens when this limit is reached.
Here, the per-response update is \(\ell_i=\sum_t m_{\epsilon^-,\epsilon^+}(r_{i,t})\hat A_i(R_\Pi(o_i;B))\).
The budget enters the reward \(R_\Pi(o_i;B)\) through the score assigned to complete and truncated responses, the normalization \(c_B(|o_i|)\) through the weight given to responses of different lengths, and the prompt-selection term \(w_B\) through which sampled groups are retained for training.
The token-level likelihood ratio \(r_{i,t}\) is transformed by the clipping rule \(m_{\epsilon^-,\epsilon^+}\), while \(\beta\) and the reference policy determine how strongly the update remains anchored to the base model.
Finally, the dataset \(\mathcal{D}\) determines how often the cap binds in the first place.
In our standard GRPO run, Qwen3.5-4B is trained on \(\mathcal{D}=\) DeepScaleR with \(B=8\)k, \(R_\Pi\) assigns \(1\) to a correct completed response and \(0\) to an incorrect or truncated one, \(\hat A_i\) standardizes this reward within a group of \(G=8\) rollouts, \(c_B(|o_i|)=1/(G|o_i|)\), \(w_B=1\) so that every group is retained, \(m_{\epsilon^-,\epsilon^+}\) is PPO clipping with \(\epsilon^-=\epsilon^+=0.2\), and \(\beta=0\).
Each ablation below changes exactly one of these factors and keeps everything else, including the optimizer, update count, and evaluation protocol, identical to this run.
For these ablations, we report avg@16 and compute every accuracy delta relative to the base model at the same evaluation budget.


\paragraph{\hypothesisquestion{Does high truncation exposure on hard datasets cause budget internalization?}}
Beyond the RL algorithm, budget internalization depends heavily on the dataset's difficulty relative to the token cap.
When a dataset requires long reasoning, many rollouts hit the cap and receive zero reward, creating a strong negative gradient against extended thinking.
To isolate this data effect, we vary only the training dataset while holding the algorithm and 8K budget fixed.
We compare GSM8K~\citep{cobbe2021gsm8k}, DeepScaleR~\citep{luo2025deepscaler}, and Nemotron-RL-Math-v2~\citep{du2025nemotronmathefficientlongcontextdistillation}, which span a wide range of natural reasoning lengths~(\Cref{tab:dataset-stats}).
We observe a trade-off between in-budget learning and test-time scaling.
DeepScaleR provides the largest in-budget gain but loses it above the training budget (\basegain{8K}{41.3}; \baseloss{64K}{22.3}).
Nemotron-RL-Math-v2 follows the same pattern (\basegain{8K}{34.2}; \baseloss{64K}{24.2}), while the simpler GSM8K dataset produces a smaller gain but preserves more of the base model's high-budget accuracy (\basegain{8K}{11.6}; \baseloss{64K}{6.8}).
\textbf{High truncation exposure on hard datasets causes the model to internalize the budget as a hard limit}, whereas training on simpler datasets preserves the model's ability to reason beyond the training budget.


\begin{table}[t]
\centering
\footnotesize
\setlength{\tabcolsep}{3pt}
\caption{AIME avg@16 accuracy (\%) of Qwen3.5-4B after GRPO ablations at an 8k training budget and after rejection fine-tuning.
Gray \(\pm\) values are 95\% confidence intervals across the 60 problem-level avg@16 scores.
Green and red values improve or degrade relative to the base model; bold and underlined values are best and second best in each column.}
\label{tab:ablation-dose-response-values}
\begin{tabularx}{\textwidth}{@{}ll*{6}{>{\raggedleft\arraybackslash}X}@{}}
\toprule
Ablation & Variant & 2k & 4k & 8k (train) & 16k & 32k & 64k \\
\midrule
\baserow Base model & & 0.0\civalue{0.0} & 0.3\civalue{0.6} & 9.2\civalue{6.2} & 34.2\civalue{10.8} & 65.4\civalue{10.1} & 78.8\civalue{8.5} \\
\input{figures/ablation_dose_response_table_rows.tex}
\end{tabularx}
\vspace{-20pt} 
\end{table}

\hypothesisquestion{Does the way we treat truncated rollouts cause budget internalization?}
A rollout does not always finish before reaching its token budget, at which point the training procedure has to decide what to do with it.
Standard GRPO assigns it zero reward, so we try three augmentations that treat such rollouts differently: loss masking removes it from the policy-gradient objective; DAPO applies a soft overlength penalty that decreases from \(0\) to \(-1\) over the final 512 tokens \citep{yu2025dapo}; and forced-answer training closes the truncated rollout's reasoning
and forces an answer prefix, giving the model up to 120 extra tokens to answer, similarly to the budget-forcing control in Section~\ref{budget-forcing-desc}.
These choices produce substantially different outcomes.
Standard GRPO produces a large in-budget gain but falls below the base model at high budget (\basegain{8K}{41.3}; \baseloss{64K}{22.3}).
DAPO worsens this trade-off (\basegain{8K}{35.0}; \baseloss{64K}{28.8}), while loss masking also does not help (\basegain{8K}{29.2}; \baseloss{64K}{19.1}).
Forced-answer training nearly removes the high-budget gap but gives up the in-budget gain (\baseloss{8K}{2.2}; \baseloss{64K}{1.6}).
Thus, \textbf{how truncated rollouts are scored has a direct effect on budget internalization}: penalizing them makes it stronger, masking them is insufficient, and recovering an answer can nearly remove it at the cost of in-budget learning.
This also suggests that the effect depends on the response-length distribution seen during training---in particular, on whether short or long responses provide the successful training signal.

\hypothesisquestion{Does response-level length normalization cause budget internalization?}
Standard GRPO averages token losses within each response using the \(1/|o_i|\) factor, where $o_i$ is the completion.
This normalization introduces a length bias: it amplifies positive updates on short correct responses and dilutes negative updates on long incorrect responses \citep{liu2025drgrpo}.
Because budget internalization reduces long-horizon reasoning, this bias is a plausible algorithmic mechanism.
We test this by rerunning the 8K-budget Qwen experiment with Dr.GRPO~\citep{liu2025drgrpo}, changing only the normalization from response length \(|o_i|\) to the fixed generation budget \(B\), while keeping the data, reward, optimizer, and update count fixed.
Dr.GRPO (\basegain{8K}{39.3}; \baseloss{64K}{23.1}) performs similarly to standard GRPO (\basegain{8K}{41.3}; \baseloss{64K}{22.3}).
\textbf{Response-level length bias is not the cause of budget internalization} in our setting.

\hypothesisquestion{Can anchoring the policy to the base model preserve elasticity?}
Budget internalization might emerge because the RL update allows the policy to drift too far from its capable initialization.
Because the base model already possesses significant long-budget reasoning capabilities, anchoring the trained policy to the base model via a Kullback-Leibler (KL) penalty, which keeps the policy closer to the reference as the penalty grows~\citep{mroueh2025reinforcementlearningverifiablerewards}, might prevent the model from forgetting how to extrapolate.
We evaluate KL regularization (\(\beta=0.001\) and \(\beta=0.01\)) while holding the dataset and 8K training budget fixed.
The weak penalty performs similarly to standard GRPO (\(\beta=0.001\): \basegain{8K}{37.9}; \baseloss{64K}{22.5}).
Increasing the penalty preserves more high-budget accuracy but reduces the in-budget gain (\(\beta=0.01\): \basegain{8K}{23.2}; \baseloss{64K}{8.6}).
\textbf{KL regularization serves as a highly effective control for preserving above-budget capability} at the cost of in-budget learning.

\hypothesisquestion{Is budget internalization an artifact of RL?}
The preceding results suggest that budget internalization depends on the response-length distribution and how truncated rollouts are penalized in RL.
However, the dynamics of GRPO might also play a key role and be critical for this effect to emerge.
To isolate the influence of the training data, we ablate the training method.
We test supervised fine-tuning (SFT) on the model's own successful rollouts (rejection fine-tuning, RFT), keeping only those that terminate naturally within 4K tokens.
We then compare this to artificially shortened traces of the same length constructed via budget forcing (\Cref{budget-forcing-desc}). Specifically, we take rollouts that naturally exceed 4K tokens, truncate their reasoning at 4096 tokens, prompt the model for its final answer with the budget-forcing mechanism, and keep only the trajectories whose forced answer matches the ground truth.
Both sets of traces are at most 4K tokens long and differ only in whether the reasoning ends naturally.
To match training exposure, both runs use the same hyperparameters and the same number of optimization steps.
Short-trace RFT does not produce the elasticity drop seen in GRPO and exceeds the base model at high budget despite no in-budget gain (\baseloss{8K}{1.4}; \basegain{64K}{4.3}), whereas budget-forced RFT reproduces the GRPO failure (\basegain{8K}{9.3}; \baseloss{64K}{38.4}).
Budget internalization is therefore not specific to RL: supervised training induces it as well, but only when the training traces contain truncation.
Naturally terminated traces reinforce concise reasoning where the model already concludes within the budget, without imposing the budget as a stopping rule, which parallels the GSM8K results, where few rollouts reach the cap.
Avoiding truncation thus mitigates budget internalization, but yields no improvement at 4K or 8K, unlike standard GRPO (\basegain{8K}{41.3}).
Overall, \textbf{budget internalization emerges whenever the training signal explicitly enforces a budget constraint} (whether through truncation penalties on hard datasets in RL, or artificially shortened trajectories in SFT).


Taken together, the ablations point away from a single GRPO-specific objective term and toward the distribution of responses reinforced during training.
Changing response-length normalization or applying an overlength penalty does not restore scaling, while reducing truncation exposure preserves substantially more high-budget accuracy.
The way truncated rollouts are scored has the largest effect: penalizing them strengthens internalization, masking them is insufficient, and recovering an answer nearly removes the high-budget gap only by giving up the in-budget gain.
KL regularization exposes the same trade-off more gradually, preserving more of the base model's high-budget capability as anchoring increases but reducing accuracy at the training budget.
The SFT controls show that this behavior is not specific to RL, since naturally completed trajectories preserve scaling whereas artificially shortened trajectories reproduce the contraction.
Across these interventions, stronger in-budget gains generally coincide with greater budget internalization, and none preserves both the full low-budget improvement and the base model's test-time scaling.
Table~\ref{tab:ablation-dose-response-values} reports the exact comparison, and Appendix Figure~\ref{fig:ablation-dose-response} shows the complete response curves.


%% file: figures/ablation_dose_response_table_rows.tex
GRPO & DeepScaleR & \bettervalue{1.1}\civalue{1.7} & \secondvalue{\bettervalue{22.7}}\civalue{9.8} & \bestvalue{\bettervalue{50.4}}\civalue{10.4} & \bettervalue{56.5}\civalue{10.0} & \worsevalue{56.5}\civalue{10.0} & \worsevalue{56.5}\civalue{10.0} \\
\midrule
Data & GSM8K & 0.0\civalue{0.0} & \bettervalue{2.8}\civalue{2.7} & \bettervalue{20.7}\civalue{9.2} & \bettervalue{52.7}\civalue{11.2} & \bestvalue{\bettervalue{70.6}}\civalue{9.4} & \worsevalue{72.0}\civalue{9.2} \\
 & Nemotron-RL-Math-v2 & \secondvalue{\bettervalue{3.4}}\civalue{2.5} & \bettervalue{19.9}\civalue{8.0} & \bettervalue{43.3}\civalue{10.1} & \bettervalue{54.0}\civalue{9.4} & \worsevalue{54.6}\civalue{9.5} & \worsevalue{54.6}\civalue{9.5} \\
\midrule
Truncation & Loss masking & \bettervalue{1.0}\civalue{1.0} & \bettervalue{16.1}\civalue{7.6} & \bettervalue{38.3}\civalue{10.7} & \secondvalue{\bettervalue{56.8}}\civalue{9.9} & \worsevalue{59.7}\civalue{9.4} & \worsevalue{59.7}\civalue{9.4} \\
 & DAPO overlength penalty & \bestvalue{\bettervalue{6.4}}\civalue{4.2} & \bestvalue{\bettervalue{27.2}}\civalue{9.9} & \bettervalue{44.2}\civalue{10.5} & \bettervalue{49.7}\civalue{10.2} & \worsevalue{50.0}\civalue{10.2} & \worsevalue{50.0}\civalue{10.2} \\
 & Forced answer & 0.0\civalue{0.0} & \worsevalue{0.1}\civalue{0.2} & \worsevalue{7.0}\civalue{5.3} & \worsevalue{31.6}\civalue{10.4} & \bettervalue{66.1}\civalue{9.8} & \secondvalue{\worsevalue{77.2}}\civalue{8.5} \\
\midrule
Normalization & Dr.GRPO & \bettervalue{1.5}\civalue{2.1} & \bettervalue{21.9}\civalue{9.5} & \secondvalue{\bettervalue{48.4}}\civalue{10.9} & \bettervalue{55.6}\civalue{10.5} & \worsevalue{55.6}\civalue{10.5} & \worsevalue{55.6}\civalue{10.5} \\
\midrule
KL anchoring & \(\beta=0.001\) & \bettervalue{0.6}\civalue{0.9} & \bettervalue{18.3}\civalue{8.6} & \bettervalue{47.1}\civalue{10.7} & \bettervalue{56.2}\civalue{10.1} & \worsevalue{56.2}\civalue{10.1} & \worsevalue{56.2}\civalue{10.1} \\
 & \(\beta=0.01\) & 0.0\civalue{0.0} & \bettervalue{8.9}\civalue{5.9} & \bettervalue{32.4}\civalue{11.0} & \bestvalue{\bettervalue{57.5}}\civalue{10.5} & \secondvalue{\bettervalue{69.4}}\civalue{8.8} & \worsevalue{70.1}\civalue{8.7} \\
\midrule
RFT & Short traces (\(<\)4K) & 0.0\civalue{0.0} & 0.3\civalue{0.6} & \worsevalue{7.8}\civalue{4.8} & \worsevalue{30.3}\civalue{10.5} & \worsevalue{63.5}\civalue{10.5} & \bestvalue{\bettervalue{83.0}}\civalue{7.9} \\
 & Budget-forced traces (4K) & 0.0\civalue{0.0} & \bettervalue{7.1}\civalue{3.1} & \bettervalue{18.5}\civalue{7.3} & \worsevalue{26.1}\civalue{9.2} & \worsevalue{37.7}\civalue{8.9} & \worsevalue{40.4}\civalue{8.5} \\
\bottomrule

%% file: literature_review.tex
\section{Related Work}
\label{sec:related}

\paragraph{Reasoning in LLMs.}
Reasoning has become a central capability of recent large language models.
It was first elicited through chain-of-thought (CoT) prompting \citep{wei2022chain, kojima2022large}, and theoretical work shows that some problems cannot be solved without such intermediate computation \citep{feng2023cot, merrill2024expressive, li2024serial}.
Subsequent work moved from eliciting reasoning at inference time to training it directly.
\citet{deepseek2025r1} showed that reinforcement learning with verifiable rewards (RLVR) alone can induce extended reasoning without human demonstrations, and current models typically combine supervised bootstrapping on externally generated traces with a subsequent RL stage \citep{qwen3technicalreport, nvidia_nemotron_3_ultra_2026}.
This RL stage runs under a fixed generation budget, and it is the setting in which we study budget internalization.
Some recent models additionally expose a reasoning-effort setting that selects the amount of reasoning at inference time \citep{qwen38, deepseekai2026deepseekv41flash}.
Such explicit budget conditioning may mitigate budget internalization by turning the training cap from an implicit prior into a clearly defined condition.

\paragraph{Limits of test-time reasoning.}
The benefits of reasoning are uneven.
Gains concentrate on mathematical and symbolic tasks and are weaker or negative elsewhere \citep{sprague2025cot, liu2025mindstep}, and even where reasoning helps, performance remains brittle under counterfactual or perturbed problems \citep{wu2024counterfactual, mirzadeh2025gsmsymbolic, dziri2023faith}.
Longer reasoning is also not uniformly better: models overthink, and accuracy can plateau or decline as the budget grows \citep{ma2025nothinking, gema2025inverse, zhao2026testtime, shojaee2025illusion}.
Our budget-forcing experiments are consistent with this picture, as base models often reach a correct answer well before they stop reasoning.
Budget internalization adds a complementary failure mode, in which training removes the model's ability to benefit from additional sequential compute.

\paragraph{Efficient and budget-aware reasoning.}
Overthinking has motivated a large body of methods that reduce unnecessary inference.
Some allocate compute adaptively by deciding when to think, estimating a token budget, or conditioning on a user-specified budget \citep{damani2024learning, zhang2025adaptthink, li2026selfbudgeteradaptivetokenallocation, wen2025budgetthinkerempoweringbudgetawarellm, xu2025elastic}, while others directly optimize reasoning length through length-controlled training, pruning, anytime objectives, or adaptive length penalties \citep{aggarwal2025l1, hou2025thinkprune, qi2025anytime, liu2025learnreasonefficientlyadaptive, xiang2025alp, yi2025shorterbetter}.
These approaches shorten reasoning by design.
We instead show that standard fixed-budget RLVR shortens it as a side effect, and that this compression transfers beyond the training domain.
The effect can be desirable when additional tokens bring little gain, as on LiveCodeBench, but harmful when they do, as on AIME.
Related studies examine how the training horizon should relate to the base policy's solution length \citep{wachi2026relative, xue2026beyond} and how exploration can let test-time compute extrapolate beyond the training budget \citep{setlur2025e3}.
We complement them by isolating which components of the training signal cause budget internalization.

%% file: conclusion.tex
\vspace{-\baselineskip}
\section{Limitations and Future Work}
\label{sec:limitations}

Our experiments show that fixed-budget post-training can change test-time scaling, but they do not establish how universal this effect is.
The tested models have four to five billion active parameters, and most experiments use mathematical reasoning; larger scales and different reasoning tasks remain to be tested.
Only the main DeepScaleR/AIME comparison uses three training seeds and both selected and final checkpoints; other runs have one training seed.
Algorithmic generality is also unresolved because our RL experiments focus on GRPO and its variants.
However, the signals implicated by our results are not specific to GRPO: any RL algorithm must decide how to score truncated trajectories, whether to anchor the policy with KL regularization, and which response lengths enter through the training data.
We therefore expect budget internalization to depend as much on these shared choices as on the optimizer itself, but direct comparisons with other RL algorithms are needed to test this hypothesis.
Our behavioral measurements also cannot determine whether the model represents the cap explicitly or whether shorter successful trajectories simply shift its response distribution.
The ablations suggest that changing a single objective term is insufficient, while changing which response lengths receive a successful training signal has a larger effect.
Training across multiple rollout budgets, conditioning explicitly on available inference compute, and retaining long successful responses are promising ways to provide that signal.
The central open problem is whether such training can preserve the low-budget gains of fixed-cap RLVR without removing the model's ability to benefit from more compute, which we leave to future work.

\vspace{-\baselineskip}
\section{Conclusion}

In this work, we introduce the term \emph{budget internalization}: a learned constraint on test-time scaling induced by fixed-budget training. Across three model families and multiple benchmarks, we demonstrate that accuracy improves near the cap while response length and performance saturate beyond it, effectively damaging the model’s capacity for test-time scaling.
No tested method removes this trade-off, and the largest changes come from the training-data distribution and the treatment of truncated rollouts.
Evaluation across reasoning budgets is therefore essential, since evaluation only near the training cap can miss a substantial loss of test-time scaling.
At the same time, budget internalization is not always harmful.
On AIME, it limits the model's ability to solve problems that require longer reasoning.
On GPQA and LiveCodeBench, however, fixed-budget checkpoints often perform better with shorter responses, showing that internalization can improve inference efficiency.
Budget forcing further shows that even base models can often produce an answer earlier without completing their natural reasoning trace, suggesting that overthinking is itself an important problem.
This efficiency effect could be deliberately exploited on tasks that benefit little from extended reasoning, provided that accuracy and output-length curves are used to separate useful compression from lost capability.

%% file: appendix.tex
\section{Training and Evaluation Details}
\label{app:setup}

Table~\ref{tab:training-config} reports the hyperparameters shared by the headline DeepScaleR and Reasoning Gym GRPO runs.
All runs use 32 prompts per step, eight generations per prompt, a global rollout batch of 256, a micro-batch size of one, three epochs, and at most 300 optimizer steps.
We use a 10-step linear warmup followed by cosine decay, bfloat16 precision, GRPO ratio clipping at \(0.2\), no reference-policy KL penalty, and seeds 42, 43, and 44 for the main DeepScaleR/AIME replications; all other training conditions use seed 42.
The rollout sampler uses temperature 1.0 and top-\(p\) 1.0.
Each generated-token cap reserves an additional 2,048 tokens for the input.
We validate every 10 steps and select the checkpoint with the highest validation accuracy.

\begin{table}[ht]
\centering
\small
\resizebox{\textwidth}{!}{
\begin{tabular}{lcccccc}
\toprule
Model & Training caps & Peak LR & Minimum LR & Optimizer & Weight decay & Training backend \\
\midrule
Qwen3.5-4B & 4k, 8k & \(5\times10^{-7}\) & \(5\times10^{-8}\) & Adam & 0.01 & Megatron \\
Gemma4-E2B & 1k, 2k & \(3\times10^{-6}\) & \(3\times10^{-7}\) & FusedAdam & 0.1 & FSDP2 \\
Nemotron-3-Nano-4B & 1k, 2k & \(2\times10^{-6}\) & \(2\times10^{-7}\) & FusedAdam & 0.1 & FSDP2 \\
\bottomrule
\end{tabular}
}
\caption{Hyperparameters for the headline fixed-budget GRPO runs.
Settings not shown in the table are shared across model families as described above.}
\label{tab:training-config}
\end{table}

For evaluation, we follow each model's Hugging Face model-card recommendation.
Table~\ref{tab:evaluation-config} lists the sampling parameters.
All evaluations enable thinking, use sampling, draw multiple generations per problem, and set the generated-token limit to the evaluation budget.
We explicitly set only temperature and top-\(k\); all other sampling parameters use the defaults from the corresponding model card.

\begin{table}[ht]
\centering
\small
\begin{tabular}{lcc}
\toprule
Model & Temperature & Top-\(k\) \\
\midrule
Qwen3.5-4B & 1.0 & 20 \\
Gemma4-E2B & 1.0 & 64 \\
Nemotron-3-Nano-4B & 1.0 & Model-card default \\
\bottomrule
\end{tabular}
\caption{Evaluation sampling hyperparameters from the corresponding Hugging Face model cards \citep{qwen3.5,gemmateam2026gemma4,nvidia2026nemotron3nano}.
For Qwen, we use the recommended reasoning parameters across all benchmarks.}
\label{tab:evaluation-config}
\end{table}

The reported token budget counts generated tokens and excludes prompt tokens.
Truncated generations receive zero reward during training and zero accuracy credit during evaluation.
Task-specific parsers and verifiers determine whether a completed final answer is correct.

\begin{figure}[t]
\centering
\includegraphics[width=\textwidth]{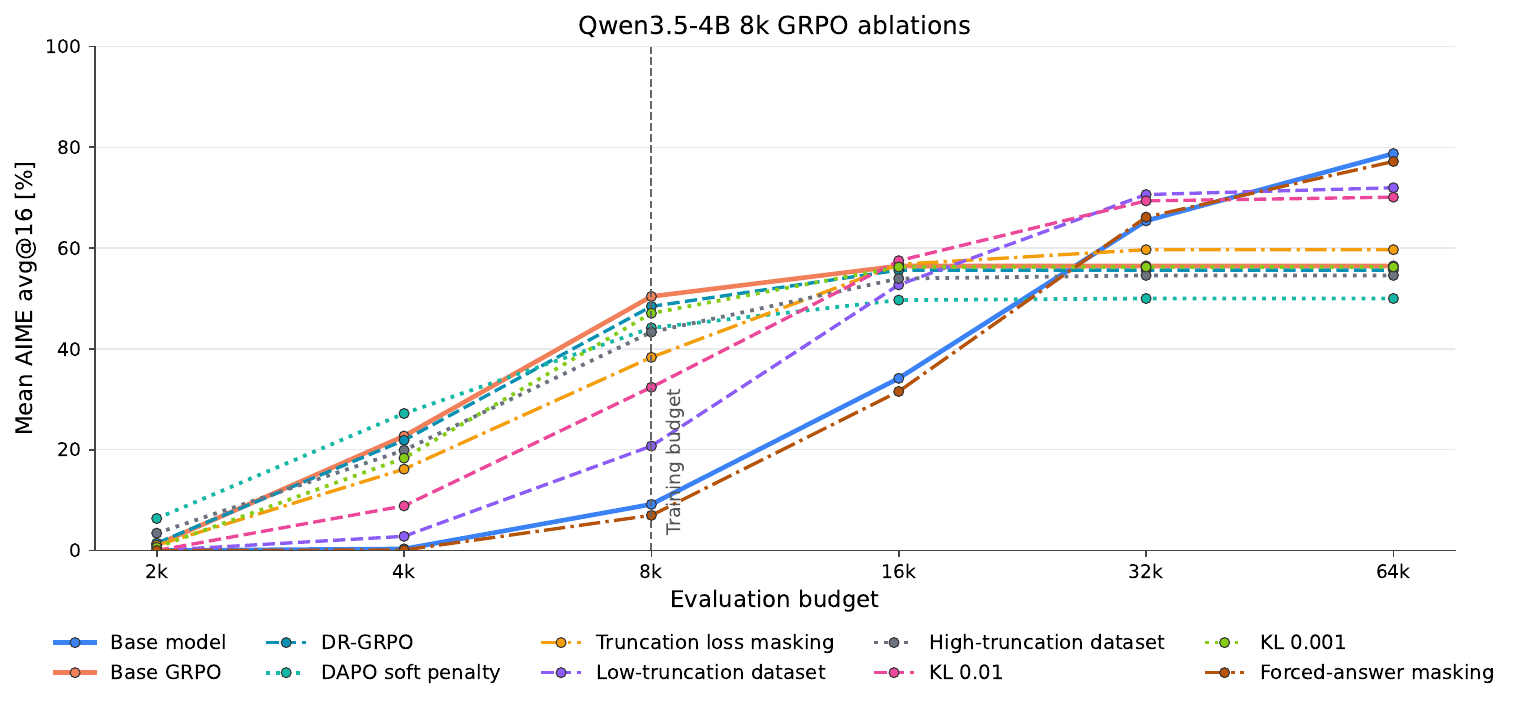}
\caption{AIME avg@16 response curves for the Qwen3.5-4B 8k training ablations.
The plot includes the base model, the standard 8k GRPO run, and all eight completed ablations.
The two KL runs use \(\beta=0.001\) and \(\beta=0.01\).
The vertical line marks the 8k training budget.
Table~\ref{tab:ablation-dose-response-values} reports the exact values.}
\label{fig:ablation-dose-response}
\end{figure}

\begin{table}[ht]
\centering
\caption{Difficulty and length statistics for the math training datasets, measured with Qwen3.5-4B in thinking mode.
Truncation at 8K is the share of responses that do not finish within 8K tokens.}
\label{tab:dataset-stats}
\begin{tabular*}{\textwidth}{@{\extracolsep{\fill}}lccc@{}}
\toprule
\textbf{Dataset} & \textbf{Pass Rate at 64K} & \textbf{Truncation at 8K} & \textbf{Median Output Tokens} \\
\midrule
GSM8K & 92\% & 24\% & 3,744 \\
DeepScaleR & 77\% & 79\% & 19,056 \\
Nemotron-RL-Math-v2 & 31\% & 99\% & 60,879 \\
\bottomrule
\end{tabular*}
\end{table}

\FloatBarrier
\section{Weak Budget Conditioning}
\label{sec:conditioning}

We test whether a stated numeric budget can control generation without changing the available compute.
In the budget-cue condition, we insert \texttt{You have a \(N\)-token reasoning budget.} immediately after the opening \texttt{<think>} tag.
The no-cue condition leaves the prompt unchanged.
Both conditions use the same 64k generation cap, while the cue states either 4k or 16k tokens.
The checkpoints trained with these cues receive the same modified prompt during evaluation and are mechanically truncated at the stated budget.
These 4k and 16k cue-conditioned checkpoints are separate training runs from the headline fixed-budget checkpoints, and the base-model control was rerun as part of this experiment.
The resulting 66.7\% no-cue base accuracy at 32k in Figure~\ref{fig:budget-cue-base-placement} therefore differs from the 64.4\% headline value because it comes from an independent evaluation sample.

\begin{figure}[t]
\centering
\begin{minipage}[t]{0.49\textwidth}
\centering
\includegraphics[width=\linewidth]{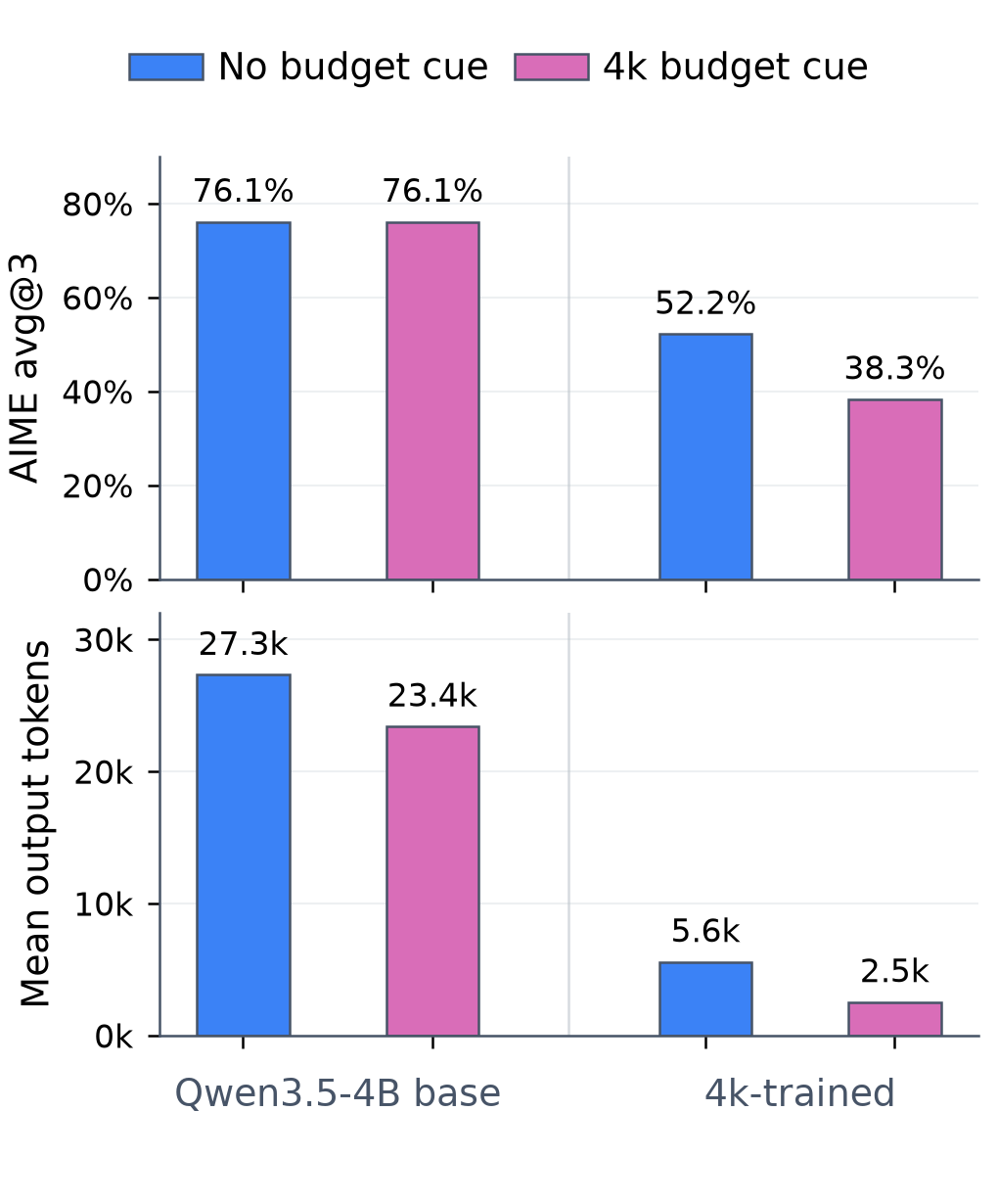}
\end{minipage}\hfill
\begin{minipage}[t]{0.49\textwidth}
\centering
\includegraphics[width=\linewidth]{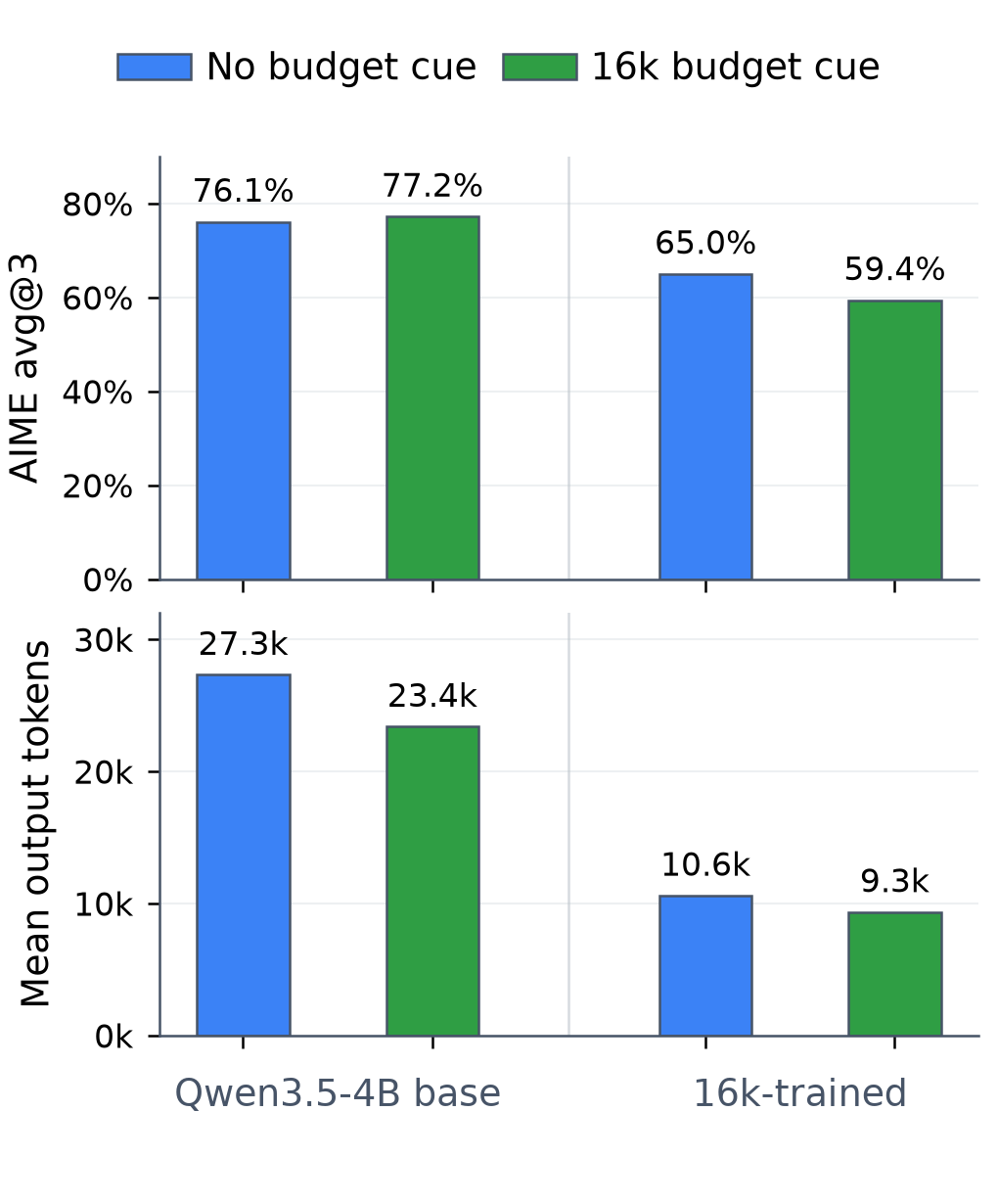}
\end{minipage}
\caption{Budget-cue ablation under a shared 64k generation cap.
Left: no cue versus a 4k cue for the base model and 4k cue-conditioned checkpoint.
Right: no cue versus a 16k cue for the base model and 16k cue-conditioned checkpoint.
The stated budget changes, but the available generation budget does not.}
\label{fig:budget-disclaimer-final}
\end{figure}

The results are presented in Figure \ref{fig:budget-disclaimer-final}. The base model does not follow either stated budget as a hard limit.
The 4k cue leaves its AIME accuracy unchanged at \(76.1\%\) and shortens its mean response only from 27.3k to 23.4k tokens.
For the 4k cue-conditioned checkpoint, the cue shortens the mean response from 5.6k to 2.5k tokens but also reduces accuracy from \(52.2\%\) to \(38.3\%\).
The 16k cue has a milder effect: the base model changes from \(76.1\%\) to \(77.2\%\) while again averaging 23.4k tokens, and the 16k cue-conditioned checkpoint shortens from 10.6k to 9.3k while accuracy falls from \(65.0\%\) to \(59.4\%\).
Thus, a cue can shift length, particularly for a checkpoint trained with that cue, but it does not provide calibrated budget control and can reduce accuracy.

\subsection{Cue Placement Ablations}
\label{app:budget-cue-placement}

\begin{figure}[h]
\centering
\includegraphics[width=\textwidth]{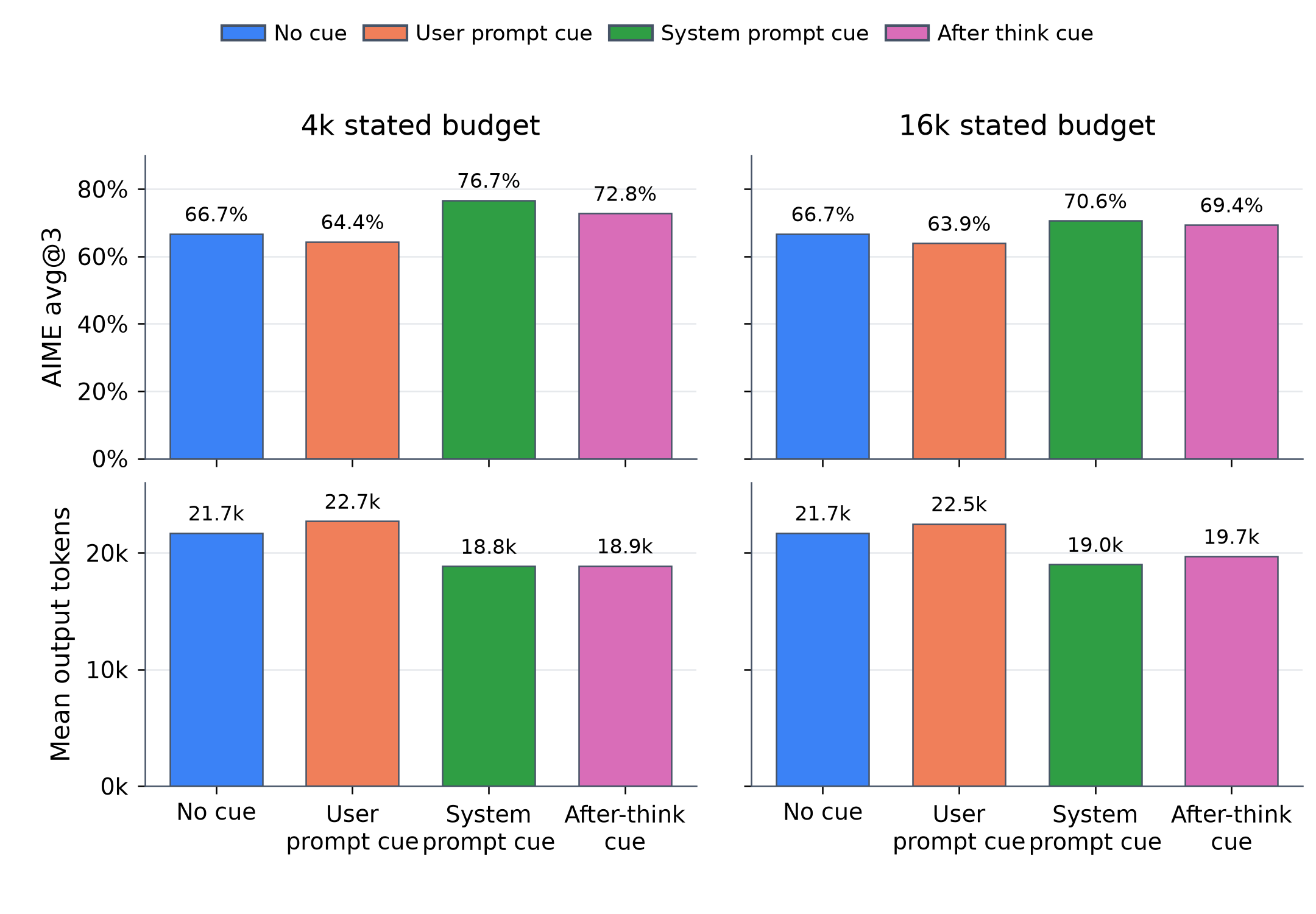}
\caption{Base-model placement ablation at a shared 32k generation cap. Columns
show 4k- and 16k-token budget cues; rows show combined AIME avg@3 and mean
output tokens. The no-cue control is shared across cue placements.}
\label{fig:budget-cue-base-placement}
\end{figure}

The base-model ablation varies whether the cue is placed in the user prompt, system prompt, or immediately after the opening \texttt{<think>}
tag. We evaluate no-cue and stated-budget-cue conditions at a shared 32k-token
generation cap. The prompt states that the reasoning budget is either 4k or 16k tokens. We report these results in Figure \ref{fig:budget-cue-base-placement}.

\begin{figure}[h]
\centering
\includegraphics[width=\textwidth]{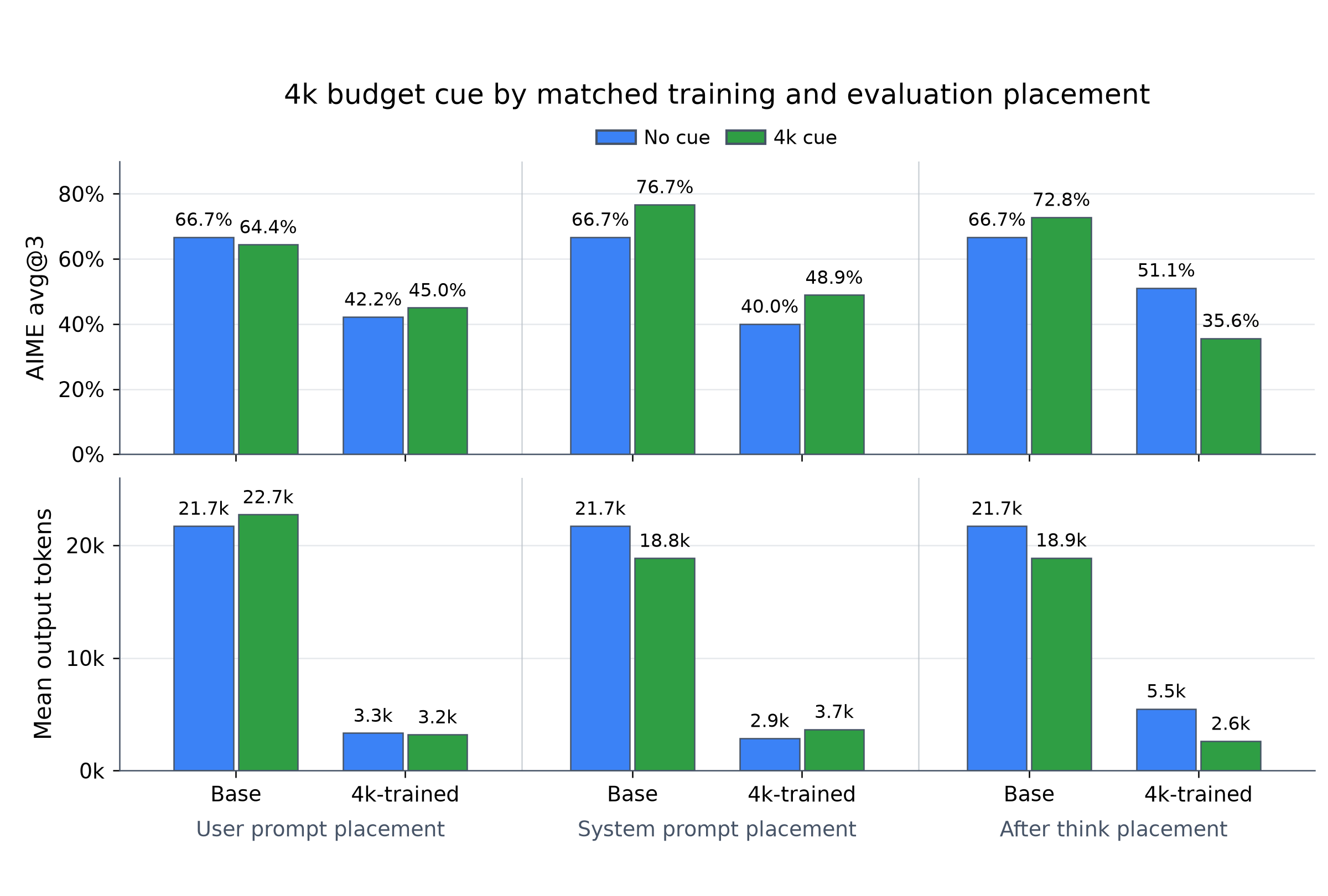}
\caption{Matched-placement ablation for separately trained 4k cue-conditioned checkpoints at a shared 32k
generation cap. Within each placement, the base model and the checkpoint trained
with that placement are evaluated with no cue (blue) or a 4k budget cue (green).
Rows show combined AIME avg@3 and mean output tokens.}
\label{fig:budget-cue-trained-placement}
\end{figure}

Additionally, we train the base model with three listed budget cue placement variants at 4k reasoning cap (the prompt during training is faithful, meaning that it always states that the reasoning budget is 4k tokens). We evaluate with
the cue in the same location used during training, also stating that the available budget is 4k, while the real truncation happens only at 32k. The results are displayed in Figure \ref{fig:budget-cue-trained-placement}. 

We observe that these alternative budget cue placements have a marginal effect on both base and trained models' output length and accuracy.

\section{Prompt and Termination Protocols}
\label{app:prompts}

\subsection{Budget-Forcing Prompt Presets}
\label{app:budget-forcing-prompts}

We match each budget-forcing preset to the model's chat template and the domain's answer format.
Figures~\ref{fig:prompt-math-qwen}--\ref{fig:prompt-lcb-nemotron} report the complete rendered prompt for every preset.
For each of them, we apply the same procedure: take a truncated model answer, insert the end-of-thinking token and benchmark-specific prefill, and allow the model a short response window for completing the answer. 

For AIME, we allow 120 tokens for continuation and apply the standard last-5-lines scorer (meaning that we look for a parseable answer in the last five lines of the response).
For GPQA, we allow 16 tokens and stop once the model has generated a newline. 
For LiveCodeBench v6, we allow 1024 additional tokens and attempt to extract a fenced Python block from the answer.

\newcommand{\crossdomainpromptfigure}[4]{
\begin{figure}[h]
\centering
\fbox{\parbox{0.92\textwidth}{
\raggedright\ttfamily
SYSTEM MESSAGE\par
\fillin\par\medskip
USER MESSAGE WITH #3\par
\fillin\par\medskip
ASSISTANT TEMPLATE AND REASONING PREFIX\par
\fillin\par\medskip
FORCING INSERTION\par
\fillin\par\medskip
RESERVED ANSWER TOKENS: \fillin\quad STOP STRINGS: \fillin\quad PARSED ANSWER: \fillin
}}
\caption{Complete #2 budget-forcing prompt for #1.}
\label{#4}
\end{figure}
}

\definecolor{promptblue}{HTML}{3B82F6}
\definecolor{promptpink}{HTML}{D96DB8}
\definecolor{promptink}{HTML}{273244}
\definecolor{promptborder}{HTML}{D8DEE9}
\newcommand{\promptsection}[1]{\par\smallskip\noindent\textbf{\textsc{#1}}\par}
\newcommand{\promptlegend}{\noindent{\color{promptpink}\rule[0.3ex]{1.2em}{1pt}}\enspace Evaluator insertion\qquad{\color{promptblue}\rule[0.3ex]{1.2em}{1pt}}\enspace Generated continuation}

\begin{figure}[p]
\centering
{\setlength{\fboxsep}{3pt}\setlength{\fboxrule}{0.6pt}
\fcolorbox{promptborder}{white}{\begin{minipage}{0.96\textwidth}
\scriptsize\raggedright\color{promptink}
\promptsection{System}
\texttt{<|im\_start|>system}\par

\texttt{<|im\_end|>}\par
\promptsection{User}
\texttt{<|im\_start|>user}\par
Solve the following math problem step by step.\par
The last line of your response should be of the form "ANSWER: ANSWER" (without quotes) where ANSWER is the answer to the problem.\par
\par
Let N be the greatest four-digit positive integer with the property that whenever one of its digits is changed to 1, the resulting number is divisible by 7. Let Q and R be the quotient and remainder, respectively, when N is divided by 1000. Find Q+R.\par
\par
Remember to put your answer on its own line at the end in the form "ANSWER: ANSWER" (without quotes) where ANSWER is the answer to the problem, and you do not need to use a \textbackslash\{\}boxed command.\par
\texttt{<|im\_end|>}\par
\promptsection{Assistant}
\texttt{<|im\_start|>assistant}\par
\texttt{<think>}\par
Let N be the greatest four-digit positive integer such that whenever one of its digits is changed to 1, the resulting number is divisible by 7.\par
Let the decimal representation of N be $d_3 d_2 d_1 d_0$, so $N = 1000 d_3 + 100 d_2 + 10 d_1 + d_0$.\par
Since N is a four-digit integer, $d_3 \in {1, 2, ..., 9}$ and $d_2, d_1, d_0 \in {0, 1, ..., 9}$.\par
The condition states that for each position $i \in {0, 1, 2, 3}$, if we replace $d_i$ with 1, the new number must be divisible by 7.\par
Let $M_i$ denote the number obtained by changing the digit at position i to 1.\par
We are given $M\_i \equiv 0 \pmod 7$ for all i=0, 1, 2, 3.\par
\par
Let's write N in terms of its digits.\par
$d_0$: units place, coefficient 1.\par
$d_1$: tens place, coefficient 10.\par
$d_2$: hundreds place, coefficient 100.\par
$d_3$: thousands place, coefficient 1000.\par
\par
\dots
\par
Diffs:\par
$(1) - (2): (6d_3 - 6d_3) + (2d_2 - 2d_2) + (3d_1 - d_0) + (1 - 3) \equiv 0 \implies 3d_1 - d_0 - 2 \equiv 0 \implies 3d_1 - d_0 \equiv 2.$\par
My previous calc was $3d_1 - d_0 - 4 \equiv 0,$ which was using $d_0$ term in (2) having coeff 1, and 3 term in (1)?\par
Wait, Eq 2 has constant term 3? No, $d_1$ becomes 1. The coefficient is 10 -> 3.\par
In N, term is $d_1 \cdot 10 \to 3d_1$. In $M_1$, term is $1 \cdot 10 \to 3$. Constant term is $d_0 \cdot 1$.\par
So Eq 2 is $6d_3 + 2d_2 + 3 + d_0 \equiv 0$. Correct.\par
Eq 1 is $6d_3 + 2d_2 + 3d_1 + 1 \equiv 0$\par
\medskip
{\color{promptpink}\texttt{</think>}}\par
{\color{promptpink}\texttt{Final Answer:}}\par
{\color{promptpink}\texttt{\textbackslash boxed\{}}{\color{promptblue}\texttt{10\}}}\par
\promptlegend
\end{minipage}}}
\caption{Complete math budget-forced example for Qwen3.5-4B on AIME 2024. The initial reasoning call reached the 4,096-token cap and was shortened here.
Pink text is injected by the evaluator; blue text is generated by the model afterward.}
\label{fig:prompt-math-qwen}
\end{figure}

\begin{figure}[p]
\centering
{\setlength{\fboxsep}{3pt}\setlength{\fboxrule}{0.6pt}
\fcolorbox{promptborder}{white}{\begin{minipage}{0.96\textwidth}
\scriptsize\raggedright\color{promptink}
\promptsection{System}
\texttt{<bos><|turn>system}\par
\texttt{<|think|>}\par
\texttt{<turn|>}\par
\promptsection{User}
\texttt{<|turn>user}\par
Solve the following math problem step by step.
The last line of your response should be of the form ``ANSWER: \$ANSWER''
(without quotes) where \$ANSWER is the answer to the problem.\par
Every morning Aya goes for a $9$-kilometer-long walk and stops at a coffee shop
afterwards. When she walks at a constant speed of $s$ kilometers per hour, the
walk takes her 4 hours, including $t$ minutes spent in the coffee shop. When
she walks $s+2$ kilometers per hour, the walk takes her 2 hours and 24 minutes,
including $t$ minutes spent in the coffee shop. Suppose Aya walks at
$s+\frac{1}{2}$ kilometers per hour. Find the number of minutes the walk takes
her, including the $t$ minutes spent in the coffee shop.\par
Remember to put your answer on its own line at the end in the form
``ANSWER: \$ANSWER'' (without quotes) where \$ANSWER is the answer to the
problem, and you do not need to use a \texttt{\textbackslash boxed} command.
\par\medskip
\texttt{<turn|>}\par
\promptsection{Assistant}
\texttt{<|turn>model}\par
\texttt{<|channel>thought}\par
Let $D$ be the distance of the walk, $D = 9$ kilometers.
Let $s$ be the speed of walking in kilometers per hour (km/h).
Let $t$ be the time spent in the coffee shop in minutes.

We are given two scenarios.

Scenario 1:
Speed is $s$ km/h.
Time taken for the walk is 4 hours, including $t$ minutes in the coffee shop.

The time spent walking is $T_1$.
$T_1 = D / s = 9 / s$ hours.
The total time elapsed is $T_1 + t/60$ hours.
We are given that the total time is 4 hours.
$9/s + t/60 = 4$ (Equation 1)

Scenario 2:
Speed is $s+2$ km/h.
Time taken for the walk is 2 hours and 24 minutes, including $t$ minutes in the coffee shop.
2 hours and 24 minutes is $2 + 24/60 = 2 + 0.4 = 2.4$ hours.
$T_2 = D / (s+2) = 9 / (s+2)$ hours.
The total time elapsed is $T_2 + t/60$ hours.
$9/(s+2) + t/60 = 2.4$ (Equation 2)

We have a system of two equations with two variables $s$ and $t$.

Let's simplify the equations.

Equation 1:
$9/s + t/60 = 4$
$t/60 = 4 - 9/s$
$t = 60 (4 - 9/s)$
$t = 240 - 540/s$ (Equation 1')

Equation 2:
$9/(s+2) + t/60 = 2.4$
$t/60 = 2.4 - 9/(s+2)$
$t = 60 (2.4 - 9/(s+2))$
$t = 144 - 540/(s+2)$ (Equation 2')

Now we equate the expressions for $t$:
$240 - 540/s = 144 - 540/(s+2)$

Let's solve for $s$.
$240 - 144 = 540/s - 540/(s+2)$
$96 = 540 [ 1/s - 1/(s+2) ]$
$96 = 540 [(s+2 - s) / (s(s+2))]$
$96 = 540 [(2) / (s^2 + 2s)]$
$96 = 1080 / (s^2 + 2s)$

$s^2 + 2s = 1080 / 96$

Let's simplify the fraction $1080 / 96$.
$1080 / 96$. Both are divisible by 12.
$1080 / 12 = 90$
$96 / 12 = 8$
$1080 / 96 = 90 / 8$.
$90 / 8 = 45 / 4$.

$s^2 + 2s = 45/4$
$4s^2 + 8s = 45$
$4s^2 + 8s - 45 = 0$

We can solve this quadratic equation for $s$ using the quadratic formula:
$s = [-8 \pm \sqrt{8^2 - 4(4)(-45)}] / (2 \cdot 4)$
$s = [-8 \pm \sqrt{64 + 16(45)}] / 8$
$16 \times 45 = 16 \times (40 + 5) = 640 + 80 = 720$.
$s = [-8 \pm \sqrt{64 + 720}] / 8$
$s = [-8 \pm \sqrt{784}] / 8$

We need $\sqrt{784}$. $20^2 = 400$, $30^2 = 900$. It ends in 4, so the root ends\par\medskip
{\color{promptpink}\texttt{<channel|>}}\par
{\color{promptpink}\texttt{Final Answer:}}\par
{\color{promptpink}\ttfamily\textbackslash boxed\{}{\color{promptblue}\texttt{30\}}}
\promptlegend
\end{minipage}}}
\caption{Complete math budget-forced example for Gemma4-E2B on AIME 2024. The initial
reasoning call reached the 1,024-token cap and is presented in full here.
Pink text is injected by the evaluator; blue text is generated by the model afterward.}
\label{fig:prompt-math-gemma}
\end{figure}

\begin{figure}[p]
\centering
{\setlength{\fboxsep}{3pt}\setlength{\fboxrule}{0.6pt}
\fcolorbox{promptborder}{white}{\begin{minipage}{0.96\textwidth}
\scriptsize\raggedright\color{promptink}
\promptsection{System}
\texttt{<|im\_start|>system}\par

\texttt{<|im\_end|>}\par
\promptsection{User}
\texttt{<|im\_start|>user}\par
Solve the following math problem step by step.\par
The last line of your response should be of the form "ANSWER: ANSWER" (without quotes) where ANSWER is the answer to the problem.\par
\par
Let N be the greatest four-digit positive integer with the property that whenever one of its digits is changed to 1, the resulting number is divisible by 7. Let Q and R be the quotient and remainder, respectively, when N is divided by 1000. Find Q+R.\par
\par
Remember to put your answer on its own line at the end in the form "ANSWER: ANSWER" (without quotes) where ANSWER is the answer to the problem, and you do not need to use a \textbackslash\{\}boxed command.\par
\texttt{<|im\_end|>}\par
\promptsection{Assistant}
\texttt{<|im\_start|>assistant}\par
\texttt{<think>}\par
We need largest four-digit number N such that if you change any digit to 1 (presumably any single digit, any position), the resulting number is divisible by 7.\par
\par
Interpretation: For each digit position i (thousands, hundreds, tens, ones), replace that digit with 1, keep other digits same, resulting 4-digit number divisible by 7.\par
\par
We need N such that all four numbers formed by changing one digit to 1 are multiples of 7.\par
\par
We need greatest such N (largest). Then compute Q = floor(N/1000) (the thousand digit), R = N mod 1000 (last three digits). Then Q+R.\par
\par
Let's denote N = abcd (digits a,b,c,d each 0-9, a$\geq$1). Changing each digit to 1 gives:\par
\par
N\_a = 1bcd (thousands digit becomes 1) = 1000*1 + 100*b + 10*c + d = 1000 + 100b + 10c + d\par
\par
N\_b = a1cd = 1000a + 100*1 + 10c + d = 1000a + 100 + 10c + d\par
\par
N\_c = ab1d = 1000a + 100b + 10*1 + d = 1000a + 100b + 10 + d\par
\par
N\_d = abc1 = 1000a + 100b + 10c + 1\par
\par
All must be $\equiv$0 mod7.\par
\par
Thus we have congruences:\par
\par
(1) 1000 + 100b + 10c + d $\equiv$ 0 (mod7)\par
\par
(2) 1000a + 100 + 10c + d $\equiv$ 0 (mod7)\par
\par
(3) 1000a + 100b + 10 + d $\equiv$ 0 (mod7)\par
\par
(4) 1000a + 100b + 10c + 1 $\equiv$ 0 (mod7)\par
\par
\dots
\par
Now d $\equiv$ 4c mod7: if c=0 $\to$ d $\equiv$0. d digit 0-9 $\to$ d=0 or 7? but must satisfy 0-9 with mod condition maybe 0 or 7 (since 7$\equiv$0 mod7). Both okay.\par
\par
Check (3): 6a +2b + d $\equiv$4. Compute 6a$\equiv$12$\equiv$5. 2b=8$\equiv$1. So 5+1+d $\equiv$4 $\to$ 6+d $\equiv$4 $\to$ d $\equiv$4-6 $\equiv$ -2 $\equiv$5 mod7. So d $\equiv$5 mod7. But we previously got d $\equiv$0 mod7. Conflict. So a=2 not possible.\par
\par
Thus a must be 9.\par
\par
Now a=9, b=4.\par
\par
Compute (4): 6a =6*9=54 $\equiv$ 54 mod7: 7*7=49 remainder5. So 5. 2b=8$\equiv$1. So 5+1+3c $\equiv$6 $\to$ 6 +3c $\equiv$6 $\to$ 3c $\equiv$0 $\to$ c $\equiv$0 mod7 $\to$ c =0 or7.\par
\par
Now (3): 6a +2b + d $\equiv$4. 6a $\equiv$5, 2b$\equiv$1 $\to$ sum 6 + d $\equiv$4 $\to$ d $\equiv$5 (since -2 $\equiv$5). So d $\equiv$5 mod7. So d =5 or 12? digit 5 only (since 5$<$10). Also d $\equiv$? check also d $\equiv$4c mod7 from earlier.\par
\par
If c=0 $\to$ d $\equiv$0 mod7, but we need d $\equiv$5 mod7 $\to$ conflict.\par
\par
If\par
{\color{promptpink}\texttt{</think>}}\par
{\color{promptpink}\texttt{Final Answer:}}\par
{\color{promptpink}\texttt{\textbackslash boxed\{}}{\color{promptblue}\texttt{1300\}}}\par
\par
\promptlegend
\end{minipage}}}
\caption{Complete math budget-forced example for Nemotron-3-Nano-4B on AIME 2024. The initial reasoning call reached the 2,048-token cap and was shortened here. Pink text is injected by the evaluator; blue text is generated by the model afterward.}
\label{fig:prompt-math-nemotron}
\end{figure}

\begin{figure}[p]
\centering
{\setlength{\fboxsep}{3pt}\setlength{\fboxrule}{0.6pt}
\fcolorbox{promptborder}{white}{\begin{minipage}{0.96\textwidth}
\scriptsize\raggedright\color{promptink}
\promptsection{System}
\texttt{<|im\_start|>system}\par

\texttt{<|im\_end|>}\par
\promptsection{User}
\texttt{<|im\_start|>user}\par
Answer the following multiple choice question. The last line of your response should be of the following format: 'ANSWER: LETTER' (without quotes) where LETTER is one of A,B,C,D. Think step by step before answering.\par
\par
Two quantum states with energies E1 and E2 have a lifetime of $10^{-9}$ sec and $10^{-8}$ sec, respectively. We want to clearly distinguish these two energy levels. Which one of the following options could be their energy difference so that they can be clearly resolved?\par
\par
\par
A) $10^{-9} eV$\par
B) $10^{-4} eV$\par
C) $10^{-11} eV$\par
D) $10^{-8} eV$\par
\texttt{<|im\_end|>}\par
\promptsection{Assistant}
\texttt{<|im\_start|>assistant}\par
The user wants me to answer a multiple-choice question about quantum states, energy levels, and time-energy uncertainty principle.\par
\par
**1. Analyze the problem:**\par
*   Two quantum states have energies E\_1 and E\_2.\par
*   They have lifetimes $\tau_1 = 10^{-9}$ sec and $\tau_2 = 10^{-8}$ sec.\par
*   We want to "clearly distinguish" or resolve these two energy levels. This means we need to determine the minimum energy separation (resolution) required given the natural linewidth associated with their finite lifetimes.\par
*   The relevant physics principle is the Time-Energy Uncertainty Principle: $\Delta E \cdot \tau \geq \hbar/2$ (or sometimes approximated as $\Delta E \approx \hbar / \tau or \Delta E \cdot \tau \sim h$). The width of the energy level line shape (natural broadening) is directly related to the lifetime of the state. Specifically, the full width at half maximum (FWHM) for a Lorentzian distribution is often taken as $\Gamma = \hbar / \tau$ in many contexts, though strictly speaking $\Delta E \Delta t \sim \hbar$. Let's assume the criterion for resolution is that the energy difference $\Delta E_{level} = |E_1 - E_2|$ must be greater than or comparable to the sum of the uncertainties (widths) of the individual states.\par

\dots 
\par
Is there any nuance where the answer might be derived differently?\par
Some texts use $\Delta E = h / \tau$.\par
$h \approx 4.14 \times 10^{-15}$ eV$\cdot$s.\par
$\Delta E_1 \approx 4.14 \times 10^{-6}$ eV.\par
$\Delta E_2 \approx 4.14 \times 10^{-7}$ eV.\par
Still, the critical threshold is $\sim 10^{-6}$ eV.\par
Option B $(10^{-4})$ is still the only one exceeding this.\par
The order of magnitude remains consistent.\par
\par
Are there cases where "clearly resolved" means much larger than the linewidth?\par
Usually "clearly resolved" implies seeing two distinct peaks. Rayleigh criterion suggests sum of widths.\par
Even sum is $\sim 10^{-6}$ eV.\par
So $10^{-4}$ eV is very clear.\par
10\par
{\color{promptpink}\texttt{</think>}}\par
{\color{promptpink}\texttt{ANSWER:}} {\color{promptblue}\texttt{B}}\par
\promptlegend
\end{minipage}}}
\caption{Complete science budget-forced example for Qwen3.5-4B on GPQA-Diamond. The initial reasoning call reached the 4,096-token cap and was shortened here. Pink text is injected by the evaluator; blue text is generated by the model afterward.}
\label{fig:prompt-gpqa-qwen}
\end{figure}

\begin{figure}[p]
\centering
{\setlength{\fboxsep}{3pt}\setlength{\fboxrule}{0.6pt}
\fcolorbox{promptborder}{white}{\begin{minipage}{0.96\textwidth}
\scriptsize\raggedright\color{promptink}
\promptsection{System}
\texttt{<bos><|turn>system}\par
\texttt{<|think|>}\par

\texttt{<turn|>}\par
\promptsection{User}
\texttt{<|turn>user}\par
Answer the following multiple choice question. The last line of your response should be of the following format: 'ANSWER: LETTER' (without quotes) where LETTER is one of A,B,C,D. Think step by step before answering.\par
\par
Consider a system of three spins $S_1$, $S_2$, and $S_3$, each of which can take spin $+1$ or $-1$. The energy is $E=-J(S_1S_2+S_1S_3+S_2S_3)$.\par
Find the partition function $Z$ of the system ($\beta=1/(kT)$, where $k$ is the Boltzmann constant and $T$ is temperature).\par
\par
A) $Z=6e^{2J\beta}+2e^{-2J\beta}$\par
B) $Z=2e^{2J\beta}+6e^{-2J\beta}$\par
C) $Z=2e^{3J\beta}+6e^{-J\beta}$\par
D) $Z=2e^{-3J\beta}+6e^{J\beta}$\par
\texttt{<turn|>}\par
\promptsection{Assistant}
\texttt{<|turn>model}\par
\texttt{<|channel>thought}\par
The user has presented a problem asking for the partition function $Z$ of three spins $S_1$, $S_2$, and $S_3$, with interaction energy $E=-J(S_1S_2+S_1S_3+S_2S_3)$.\par
\par
The spins can take values +1 or -1.\par
\par
The partition function Z is defined as:\par
$Z=\sum_{\text{states}}e^{-\beta E}$\par
where $\beta=1/(kT)$.\par
\par
The system has $2^3=8$ possible microstates. We calculate $E$ for each state and sum the Boltzmann factors $e^{-\beta E}$.\par
\par
The energy function is $E=-J(S_1S_2+S_1S_3+S_2S_3)$.\par
\par
Let's list the possible spin configurations $(S_1, S_2, S_3)$ and calculate E.\par
\par
The term $S_1S_2 + S_1S_3 + S_2S_3$ is the sum of interaction terms. Since $S_i \in \{-1, +1\}$, the product $S_iS_j$ is either +1 (if $S_i = S_j$) or -1 (if $S_i \neq S_j$).\par
\par
Let $N_{\mathrm{same}}$ be the number of pairs (i, j) such that $S_i = S_j$.\par
$S_1S_2 = 1$ if $S_1=S_2$, -1 otherwise.\par
$S_1S_3 = 1$ if $S_1=S_3$, -1 otherwise.\par
$S_2S_3 = 1$ if $S_2=S_3$, -1 otherwise.\par
\par
$E=-J\sum_{i<j}S_iS_j$.\par
\par
We can analyze the configurations based on the number of +1 spins and -1 spins.\par
\par
Case 1: All spins are the same (3 states).\par
1a) (+1, +1, +1):\par
$S_1S_2=1$, $S_1S_3=1$, and $S_2S_3=1$.\par
$S_1S_2+S_1S_3+S_2S_3=1+1+1=3$.\par
$E=-3J$.\par
$e^{-\beta E}=e^{3J\beta}$.\par
There is 1 such state.\par
\par
1b) (-1, -1, -1):\par
$S_1S_2=1$, $S_1S_3=1$, and $S_2S_3=1$.\par
$S_1S_2 + S_1S_3 + S_2S_3 = 3$.\par
$E=-3J$.\par
$e^{-\beta E}=e^{3J\beta}$.\par
There is 1 such state.\par
\par
Total contribution from Case 1: $2e^{3J\beta}$.\par
\par
Case 2: Two spins are the same, one is different (6 states).\par
2a) (+1, +1, -1): (and permutations: (+1, -1, +1), (-1, +1, +1)).\par
For (+1, +1, -1):\par
$S_1S_2=1$, $S_1S_3=-1$, and $S_2S_3=-1$.\par
$S_1S_2+S_1S_3+S_2S_3=1-1-1=-1$.\par
$E=-J(-1)=J$.\par
$e^{-\beta E}=e^{-J\beta}$.\par
There are 3 such states with energy J.\par
Contribution: $3e^{-J\beta}$.\par
\par
2b) (-1, -1, +1): (and permutations: (-1, +1, -1), (+1, -1, -1)).\par
For (-1, -1, +1):\par
$S_1S_2=1$, $S_1S_3=-1$, and $S_2S_3=-1$.\par
$S_1S_2+S_1S_3+S_2S_3=1-1-1=-1$.\par
$E=J$.\par
$e^{-\beta E}=e^{-J\beta}$.\par
There are 3 such states\par
{\color{promptpink}\texttt{<channel|>}}\par
{\color{promptpink}\texttt{ANSWER:}} {\color{promptblue}\texttt{B}}\par
\promptlegend
\end{minipage}}}
\caption{Complete science budget-forced example for Gemma4-E2B on GPQA-Diamond. The initial reasoning call reached the 1,024-token cap and is presented in full here. Pink text is injected by the evaluator; blue text is generated by the model afterward.}
\label{fig:prompt-gpqa-gemma}
\end{figure}

\begin{figure}[p]
\centering
{\setlength{\fboxsep}{3pt}\setlength{\fboxrule}{0.6pt}
\fcolorbox{promptborder}{white}{\resizebox{0.96\textwidth}{!}{\begin{minipage}{1.07\textwidth}
\scriptsize\raggedright\color{promptink}
\promptsection{System}
\texttt{<|im\_start|>system}\par

\texttt{<|im\_end|>}\par
\promptsection{User}
\texttt{<|im\_start|>user}\par
Answer the following multiple choice question. The last line of your response should be of the following format: 'ANSWER: LETTER' (without quotes) where LETTER is one of A,B,C,D. Think step by step before answering.\par
\par
A spin-half particle is in a linear superposition 0.5|$\uparrow\rangle$+$\sqrt{3}/2$|$\downarrow\rangle$ of its spin-up and spin-down states. If |$\uparrow\rangle$ and |$\downarrow\rangle$ are the eigenstates of $\sigma_z$ , then what is the expectation value up to one decimal place, of the operator 10$\sigma_z$+5$\sigma_x$ ? Here, symbols have their usual meanings\par
\par
A) 1.65\par
B) -0.7\par
C) 0.85\par
D) -1.4\par
\texttt{<|im\_end|>}\par
\promptsection{Assistant}
\texttt{<|im\_start|>assistant}\par
\texttt{<think>}\par
We need to answer a multiple choice about expectation value of operator O = 10 $\sigma_z$ + 5 $\sigma_x$ for a spin-1/2 state $|\psi\rangle$ = 0.5 $|\uparrow\rangle$ + $\sqrt{3}/2$ $|\downarrow\rangle$, where $|\uparrow\rangle$ and $|\downarrow\rangle$ are eigenstates of $\sigma_z$.\par
\par
We are to find expectation value $\langle\psi|O|$$\psi\rangle$ up to one decimal place.\par
\par
We need to compute expectation for O = 10 $\sigma_z$ + 5 $\sigma_x$.\par
\par
We have expectation = 10 $\langle\psi|\sigma_z|\psi\rangle$ + 5 $\langle\psi|\sigma\_x|\psi\rangle$.\par
\par
First compute $\langle\psi|\sigma_z|\psi\rangle$.\par
\par
We know $\sigma_z$ $|\uparrow\rangle$ = +1 $|\uparrow\rangle$, $\sigma_z$ $|\downarrow\rangle$ = -1 $|\downarrow\rangle$.\par
\par
Hence $\langle\psi|\sigma_z|\psi\rangle = (0.5)^*\langle\uparrow| + (\sqrt{3}/2)^* \langle\downarrow| (\sigma_z) (0.5 |\uparrow\rangle + \sqrt{3}/2 |\downarrow\rangle) = ?$\par
\par
Simplify: $\sigma_z|\psi\rangle = 0.5 |\uparrow\rangle - (\sqrt{3}/2) |\downarrow\rangle$ (since $\sigma_z$ flips sign on $\downarrow$). Then expectation = $\langle\psi|\sigma_z|\psi\rangle = \langle\psi| (0.5 |\uparrow\rangle - (\sqrt{3}/2) |\downarrow\rangle) = (0.5*0.5) + ((\sqrt{3}/2)*(\sqrt{3}/2))?$ Wait careful.\par
\par
Let's compute: $\langle\psi|\sigma_z|\psi\rangle = (0.5* \langle\uparrow| + (\sqrt{3}/2)* \langle\downarrow|) (\sigma_z (0.5 |\uparrow\rangle + (\sqrt{3}/2) |\downarrow\rangle))$\par
\par
But ($\sigma_z (0.5 |\uparrow\rangle) = 0.5 |\uparrow\rangle, (\sigma_z ( (\sqrt{3}/2) |\downarrow\rangle) = - (\sqrt{3}/2) |\downarrow\rangle )$. So we have = $(0.5* \langle\uparrow|) (0.5|\uparrow\rangle) + (\sqrt{3}/2 * \langle\downarrow|) ( - (\sqrt{3}/2) |\downarrow\rangle) = 0.5*0.5 - (\sqrt{3}/2)*(\sqrt{3}/2) = 0.25 - (3/4) = 0.25 - 0.75 = -0.5$.\par
\par
Thus $\langle\psi|\sigma_z|\psi\rangle = -0.5$.\par
\par
Alternatively could compute directly using probabilities: Expectation value of $\sigma_z = (p\_up - p\_down) = (p\_up - p\_down)$ because eigenstates with eigenvalues +1 and -1. Here $p\_up = |0.5|^2 = 0.25, p\_down = | \sqrt{3}/2 |^2 = 3/4 = 0.75$. So expectation = 0.25 - 0.75 = -0.5. Yes consistent.\par
\par
Thus $10 * \langle\sigma_z\rangle = 10 * (-0.5) = -5$.\par
\par
Now compute $\langle\psi|\sigma_x|\psi\rangle$.\par
\par
$\sigma_x$ flips spin: $\sigma_x |\uparrow\rangle = |\downarrow\rangle, \sigma_x |\downarrow\rangle = |\uparrow\rangle$.\par
\par
Thus $\sigma_x |\psi\rangle = 0.5 |\downarrow\rangle + (\sqrt{3}/2) |\uparrow\rangle$.\par
\par
Thus $\langle\psi|\sigma_x|\psi\rangle = \langle\psi| (0.5 |\downarrow\rangle + (\sqrt{3}/2) |\uparrow\rangle)$. Compute:\par
\par
$\langle\psi| = 0.5 \langle\uparrow| + \sqrt{3}/2) \langle\downarrow|$.\par
\par
So expectation = $(0.5 \langle\uparrow|) (0.5 |\downarrow\rangle) + (\sqrt{3}/2\langle\downarrow|) (\sqrt{3}/2 |\uparrow\rangle) = 0.5*0 + (   \sqrt{3}/2)*(\sqrt{3}/2)\langle\downarrow|\uparrow\rangle$?? Wait need inner product: $\langle\downarrow|\uparrow\rangle = 0$. So the first term is zero because $\langle\uparrow|\downarrow\rangle = 0$. So expectation = $(\sqrt{3}/2)*(\sqrt{3}/2)*\langle\downarrow|\uparrow\rangle$? That's zero.\par
\par
Thus $\langle\psi|\sigma_x|\psi\rangle = 0$.\par
\par
Alternatively expectation value of $\sigma_x$ = $\langle$$\sigma_x$$\rangle$ = 0 if the state is real phase? Actually generic expectation depends on off-diagonal elements of $\sigma_x$ in the state basis. Since state is (a,b) with a=0.5, b=$\sqrt{3}/2$, real. Then expectation <$\sigma_x$> = 2 Im(a* b). Since a and b are both real, Im = 0. So expectation = 0. Yes.\par
\par
Thus 5 * $\langle\sigma_x\rangle$ = 0.\par
\par
Hence expectation value = -5 + 0 = -5.\par
\par
But expected answer choices: A) 1.65, B) -0.7, C) 0.85, D) -1.4. None is -5. Something off? Let's double-check the scaling: The operator is 10$\sigma_z$ + 5$\sigma_x$. Could they intend expectation per unit? Something else? Let's compute again but consider that expectation of $\sigma_z$ = (p\_up - p\_down) = -0.5. So 10 * (-0.5) = -5. Then plus 5*0 = -5. So result -5.\par
\par
But choices are much smaller magnitude. Could we have mistaken sign of $\langle\sigma_z\rangle$? Perhaps expectation = $(p\_up - p\_down) = 0.25 - 0.75 = -0.5$. Yes.\par
\par
Thus answer -5.\par
\par
But options are far off. Maybe they want expectation value per operator components: like expectation of $\sigma_z = p\_up - p\_down = -0.5$. So expectation value of 10 $\sigma_z$ is -5. Add 5 $\sigma_x$ expectation of $\sigma_x$ = (2 Im( a* b )) = 0. So final -5. But none matches.\par
\par
Let's verify calculations for $\langle\psi|\sigma_x|\psi\rangle = 2\operatorname{Re}(a^*b)$? Wait formula: $\langle\sigma_x\rangle = 2\operatorname{Im}(a^*b)$? Actually for any state $|\psi\rangle = a|\uparrow\rangle + b|\downarrow\rangle$ (real coefficients), expectation of $\sigma_x$ = $2\operatorname{Im}(a^*b)$? That's for off-diagonal in Pauli basis? Let's recall: $\sigma_x = [[0,1],[1,0]]$.\par
\par
$\langle\psi|\sigma_x|\psi\rangle = |a* b + b* a| = 2 Re(a* b)$. Actually let's compute: $\langle\psi|\sigma_x|\psi\rangle = a* (\sigma_x|\psi\rangle)$?? Actually compute:\par
\par
$|\psi\rangle = a|\uparrow\rangle + b|\downarrow\rangle$.\par
$\sigma_x|\psi\rangle = a|\downarrow\rangle + b|\uparrow\rangle$.\par
Thus $\langle\psi|\sigma_x|\psi\rangle = (a^*\langle\uparrow| + b^*\langle\downarrow|)(a|\downarrow\rangle + b|\uparrow\rangle) = a^*a \langle\uparrow|\downarrow\rangle + a^*b \langle\uparrow|\uparrow\rangle + b^*a \langle\downarrow|\uparrow\rangle + b^*b \langle\downarrow|\downarrow\rangle = 0 + a^*b + 0 + b^*b$? Wait careful:\par
\par
$\langle\uparrow|\uparrow\rangle = 1 => a* b * 1 = a* b$.\par
\par
$\langle\downarrow|\downarrow\rangle = 1 => b* b = |b|^2$.\par
\par
But also cross term $a* b \langle\uparrow|\uparrow\rangle = a* b$.\par
\par
Also $b* a \langle\downarrow|\uparrow\rangle= 0$.\par
\par
Thus expectation = $a* b + b* b$? Actually let's recalc:\par
\par
$\langle\psi|\sigma_x|\psi\rangle = (a* \langle\uparrow| + b* \langle\downarrow|)(\sigma_x|\psi\rangle) = (a* \langle\uparrow| + b* \langle\downarrow|)(a|\downarrow\rangle + b|\uparrow\rangle) = a* a \langle\uparrow|\downarrow\rangle + a* b \langle\uparrow|\uparrow\rangle + b* a \langle\downarrow|\uparrow\rangle + b* b \langle\downarrow|\downarrow\rangle$.\par
\par
Now $\langle\uparrow|$\par
{\color{promptpink}\texttt{</think>}}\par
{\color{promptpink}\texttt{ANSWER:}} {\color{promptblue}\texttt{B}}\par
\promptlegend
\end{minipage}}}}
\caption{Complete science budget-forced example for Nemotron-3-Nano-4B on GPQA-Diamond. The initial reasoning call reached the 2,048-token cap and is presented in full here. Pink text is injected by the evaluator; blue text is generated by the model afterward.}
\label{fig:prompt-gpqa-nemotron}
\end{figure}

\begin{figure}[p]
\centering
{\setlength{\fboxsep}{3pt}\setlength{\fboxrule}{0.6pt}
\fcolorbox{promptborder}{white}{\resizebox{0.96\textwidth}{!}{\begin{minipage}{1.07\textwidth}
\scriptsize\raggedright\color{promptink}
\promptsection{System}
\texttt{<|im\_start|>system}\par
You are an expert Python programmer. You will be given a question (problem specification) and will generate a correct Python program that matches the specification and passes all tests.\par
\texttt{<|im\_end|>}\par
\promptsection{User}
\texttt{<|im\_start|>user}\par
\#\#\# Question:\par
You are given an integer X not less than 2.\par
Find the positive integer N such that N! = X.\par
Here, N! denotes the factorial of N, and it is guaranteed that there is exactly one such N.\par
\par
Input\par
\par
The input is given from Standard Input in the following format:\par
X\par
\par
Output\par
\par
Print the answer.\par
\par
Constraints\par
\par
\par
- $2 \leq X \leq 3 \times 10^{18}$\par
- There is exactly one positive integer N such that N! = X.\par
- All input values are integers.\par
\dots
\par
\texttt{<|im\_end|>}\par
\promptsection{Assistant}
\texttt{<|im\_start|>assistant}\par
\texttt{<think>}\par
The user wants a Python program to solve the following problem:\par
Given an integer $X$ ($2\leq X\leq3\times10^{18}$), find a positive integer $N$ such that $N!=X$.\par
It is guaranteed that such an N exists and is unique.\par
\par
Constraints analysis:\par
$X$ can be up to $3\times10^{18}$.\par
Let's estimate the maximum value of N.\par
$10!=3{,}628{,}800\approx3.6\times10^6$.\par
$20!\approx2.4\times10^{18}$.\par
$21!\approx5.1\times10^{19}$.\par
Since $X\leq3\times10^{18}$, $N$ will not exceed $20$. Wait, checking exact values.\par
$20!=2{,}432{,}902{,}008{,}176{,}640{,}000<3\times10^{18}$.\par
$21!=51{,}090{,}942{,}171{,}709{,}440{,}000>3\times10^{18}$.\par
\par
\par
\dots
\par
Is there any interpretation where X is not an integer? "All input values are integers." OK.\par
Format: "Print the answer."\par
\par
Let's double check the sample outputs.\par
Sample 1:\par
Input: 6\par
Output: 3\par
Matches.\par
\par
Sample 2:\par
Input: 2432902008176640000\par
Output: 20\par
Matches.\par
\par
Corner cases:\par
X = 2. Loop runs once. `fact` becomes 2. `n` becomes 2. Prints 2. Correct.\par
X = 20!.\par
Suppose X = 20!.\par
Loop continues until `fact` becomes 20!.\par
Inside loop, `fact` is initially 19!. Condition `$fact < X$` holds ($19! < 20!$).\par
Body executes: `fact` becomes 20!\par
{\color{promptpink}\texttt{</think>}}\par
{\color{promptpink}\texttt{```python}}\par
{\color{promptblue}
import sys\par
\par
def solve():\par
\quad \# Read all input from standard input\par
\quad input\_data = sys.stdin.read().split()\par
    \par
\quad  not input\_data:\par
\quad\quad return\par
    \par
\quad    \# Parse the integer X\par
\quad    \# Using int() automatically handles large integers in Python\par
\quad    x\_val = int(input\_data[0])\par
    \par
\quad    n = 1\par
\quad    fact = 1\par
    \par
\quad    \# We need to find N such that N! = X.\par
\quad    \# Given $X <= 3 * 10^{18}$, N will be at most 20 because $21! > 3 * 10^{18}$.\par
\quad    \# Therefore, a simple loop calculating factorials sequentially is efficient enough.\par
    \par
\quad    while fact < x\_val:\par
\quad\quad        n += 1\par
\quad\quad        fact *= n\par
        \par
\quad    \# The problem guarantees that exactly one such positive integer N exists.\par
\quad    \# With this logic, 'fact' will eventually equal 'x\_val' and the loop terminates.\par
\quad    print(n)\par
\par
if \_\_name\_\_ == '\_\_main\_\_':\par
    solve()\par{\texttt{```}}\par
}
\promptlegend
\end{minipage}}}}
\caption{Complete code budget-forced example for Qwen3.5-4B on LiveCodeBench v6. The initial reasoning call reached the 4,096-token cap and was shortened here; full task description was also shortened. Pink text is injected by the evaluator; blue text is generated by the model afterward.}
\label{fig:prompt-lcb-qwen}
\end{figure}

\begin{figure}[p]
\centering
{\setlength{\fboxsep}{3pt}\setlength{\fboxrule}{0.6pt}
\fcolorbox{promptborder}{white}{\begin{minipage}{0.96\textwidth}
\scriptsize\raggedright\color{promptink}
\promptsection{System}
\texttt{<bos><|turn>system}\par
\texttt{<|think|>}\par
You are an expert Python programmer. You will be given a question (problem specification) and will generate a correct Python program that matches the specification and passes all tests.\par
\texttt{<turn|>}\par
\promptsection{User}
\texttt{<|turn>user}\par
\#\#\# Question:\par
You are given an integer array nums of size $n$. For each index i where $0 <= i < n$, define a subarray nums[start ... i] where start = max(0, i - nums[i]).\par
Return the total sum of all elements from the subarray defined for each index in the array.\par
 \par
Example 1:\par
\par
Input: nums = [2,3,1]\par
Output: 11\par
Explanation:\par
\dots
\par
\texttt{<turn|>}\par
\promptsection{Assistant}
\texttt{<|turn>model}\par
\texttt{<|channel>thought}\par
Thinking Process:\par
\par
1.  **Understand the Goal:** The task is to calculate the sum of elements for a specific subarray defined for every index i in the input array `nums`.\par
    For each index $i$, the subarray is \texttt{nums[start ... i]}, where $\mathrm{start}=\max(0,i-\texttt{nums}[i])$. We need the total of these subarray sums.\par
\par
2.  **Analyze the definition:**\par
    Let $n$ be the length of \texttt{nums}. We iterate $i$ from $0$ to $n-1$.\par
    For a given i:\par
    a. Calculate $\mathrm{start}_i=\max(0,i-\texttt{nums}[i])$.\par
    b. The subarray is $\texttt{nums}[\mathrm{start}_i],\ldots,\texttt{nums}[i]$.\par
    c. Calculate $S_i=\sum_{j=\mathrm{start}_i}^{i}\texttt{nums}[j]$.\par
    d. The final result is $\sum_{i=0}^{n-1}S_i$.\par
\par
3.  **Example 1 Walkthrough: `nums = [2, 3, 1]` (n=3)**\par
\par
    *   **i = 0:**\par
        nums[0] = 2.\par
        $\mathrm{start}_0=\max(0,0-2)=\max(0,-2)=0$.\par
        Subarray: \texttt{nums[0..0]} = $[2]$; $S_0=2$.\par
    *   **i = 1:**\par
        nums[1] = 3.\par
        $\mathrm{start}_1=\max(0,1-3)=\max(0,-2)=0$.\par
        Subarray: \texttt{nums[0..1]} = $[2,3]$; $S_1=5$.\par
    *   **i = 2:**\par
        nums[2] = 1.\par
        $\mathrm{start}_2=\max(0,2-1)=\max(0,1)=1$.\par
        Subarray: \texttt{nums[1..2]} = $[3,1]$; $S_2=4$.\par
    *   Total sum: $S_0+S_1+S_2=2+5+4=11$. (Matches example output.)\par
\par
\dots 
\par
5.  **Implementation Strategy (Direct Simulation):**\par
    A straightforward approach is to iterate through i. For each i, calculate start, and then calculate the sum from start to i.\par
\par
    *   N is up to 100. An $O(N^2)$ solution is perfectly acceptable.\par
    *   Outer loop for i: 0 to N-1.\par
    *   Inner calculation (finding start and summing): O(i - start + 1), which is O(\par
{\color{promptpink}\texttt{<channel|>}}\par
{\color{promptpink}\texttt{```python}}\par
{\color{promptblue}
from typing import List\par
\par
class Solution:\par
\quad    def subarraySum(self, nums: List[int]) -> int:\par
\quad        n = len(nums)\par
\quad        total\_sum = 0\par
        \par
\quad        \# Iterate through all ending indices i\par
\quad        for i in range(n):\par
\qquad            \# 1. Determine the starting index: start = max(0, i - nums[i])\par
\qquad            start = max(0, i - nums[i])\par
            \par
\qquad            \# 2. Calculate the sum of the subarray nums[start ... i]\par
\qquad            current\_subarray\_sum = 0\par
 \qquad          for j in range(start, i + 1):\par
\qquad \quad               current\_subarray\_sum += nums[j]\par
            \par
\qquad\quad           \# 3. Add this sum to the total\par
\qquad \quad            total\_sum += current\_subarray\_sum\par
            \par
\quad        return total\_sum\par{\texttt{```}}\par
}
\promptlegend
\end{minipage}}}
\caption{Complete code budget-forced example for Gemma4-E2B on LiveCodeBench v6. The initial reasoning call reached the 1,024-token cap and was shortened here; full task description was also shortened. Pink text is injected by the evaluator; blue text is generated by the model afterward.
}
\label{fig:prompt-lcb-gemma}
\end{figure}

\begin{figure}[p]
\centering
{\setlength{\fboxsep}{3pt}\setlength{\fboxrule}{0.6pt}
\fcolorbox{promptborder}{white}{\begin{minipage}{0.96\textwidth}
\scriptsize\raggedright\color{promptink}
\promptsection{System}
\texttt{<|im\_start|>system}\par
You are an expert Python programmer. You will be given a question (problem specification) and will generate a correct Python program that matches the specification and passes all tests.\par
\texttt{<|im\_end|>}\par
\promptsection{User}
\texttt{<|im\_start|>user}\par
\#\#\# Question:\par
You are given an integer X not less than 2.\par
Find the positive integer N such that N! = X.\par
Here, N! denotes the factorial of N, and it is guaranteed that there is exactly one such N.\par
\par
Input\par
\par
The input is given from Standard Input in the following format:\par
X\par
\par
Output\par
\par
Print the answer.\par
\par
Constraints\par
\par
\par
- $2\leq X\leq3\times10^{18}$\par
- There is exactly one positive integer N such that N!=X.\par
- All input values are integers.\par
\dots
\par
\texttt{<|im\_end|>}\par
\promptsection{Assistant}
\texttt{<|im\_start|>assistant}\par
\texttt{<think>}\par
Okay, let's see. The problem is: given an integer X (at least 2), find the positive integer N such that N! equals X. And it's guaranteed there's exactly one such N.\par
\par
First, I need to think about how factorials grow. N! for N starting at 1 is 1, 2, 6, 24, 120, etc. Since X is at least 2, N is at least 2. But the input X can be as large as 3e18. So, what's the maximum possible N here?\par
\par
Let me calculate some factorials to estimate. Let's see:\par
\par
1! = 1\par
2! = 2\par
3! = 6\par
4! = 24\par
5! = 120\par
6! = 720\par
7! = 5040\par
8! = 40320\par
9! = 362880\par
\dots
\par
But note: X can be as large as 3e18, but the factorial for n=20 is 2.43e18, which is within the range of Python integers (Python integers can be arbitrarily large, but for n up to 20, the factorial is manageable).\par
\par
So the code:\par
\par
Read X.\par
\par
n = 1\par
\par
fact = 1\par
\par
while $fact < X$:\par
\par
\quad n += 1\par
\par
\quad fact *= n\par
\par
print(n)\par
\par
Yes.\par
\par
Wait, but what if X is 1? But constraints say $X >=2$, so no problem.\par
\par
Another edge case: X=2 (n=2). The loop runs once: n becomes 2, fact=2, then \par
{\color{promptpink}\texttt{</think>}}\par
{\color{promptpink}\texttt{```python}}\par
{\color{promptblue}
def main():\par
\quad    X = int(input().strip())\par
\quad    n = 1\par
\quad    fact = 1\par
\quad    while $fact < X$:\par
\qquad        n += 1\par
\qquad        fact *= n\par
\quad   print(n)\par
\par
if \_\_name\_\_ == '\_\_main\_\_': \par
    main()\par
\texttt{```}\par
}
\promptlegend
\end{minipage}}}
\caption{Complete code budget-forced example for Nemotron-3-Nano-4B on LiveCodeBench v6. The initial reasoning call reached the 2,048-token cap and was shortened here; full task description was also shortened. Pink text is injected by the evaluator; blue text is generated by the model afterward.}
\label{fig:prompt-lcb-nemotron}
\end{figure}


\subsection{Extending Reasoning with \texttt{Wait}}\label{app:add-wait}

We take the 4k-trained Qwen3.5-4B checkpoint's initial generation, discard its final answer, remove the end-of-thinking delimiter, append \texttt{Wait}, and ask it to continue the same reasoning trace.
We repeat the insertion up to four times under a maximum 64k response length and evaluate on AIME24\&25, GPQA-Diamond, and LiveCodeBench v6.
For comparison, we report ordinary base and trained-checkpoint decoding together with answer forcing on the base model.

\begin{figure}[h!]
\centering
\includegraphics[width=\textwidth]{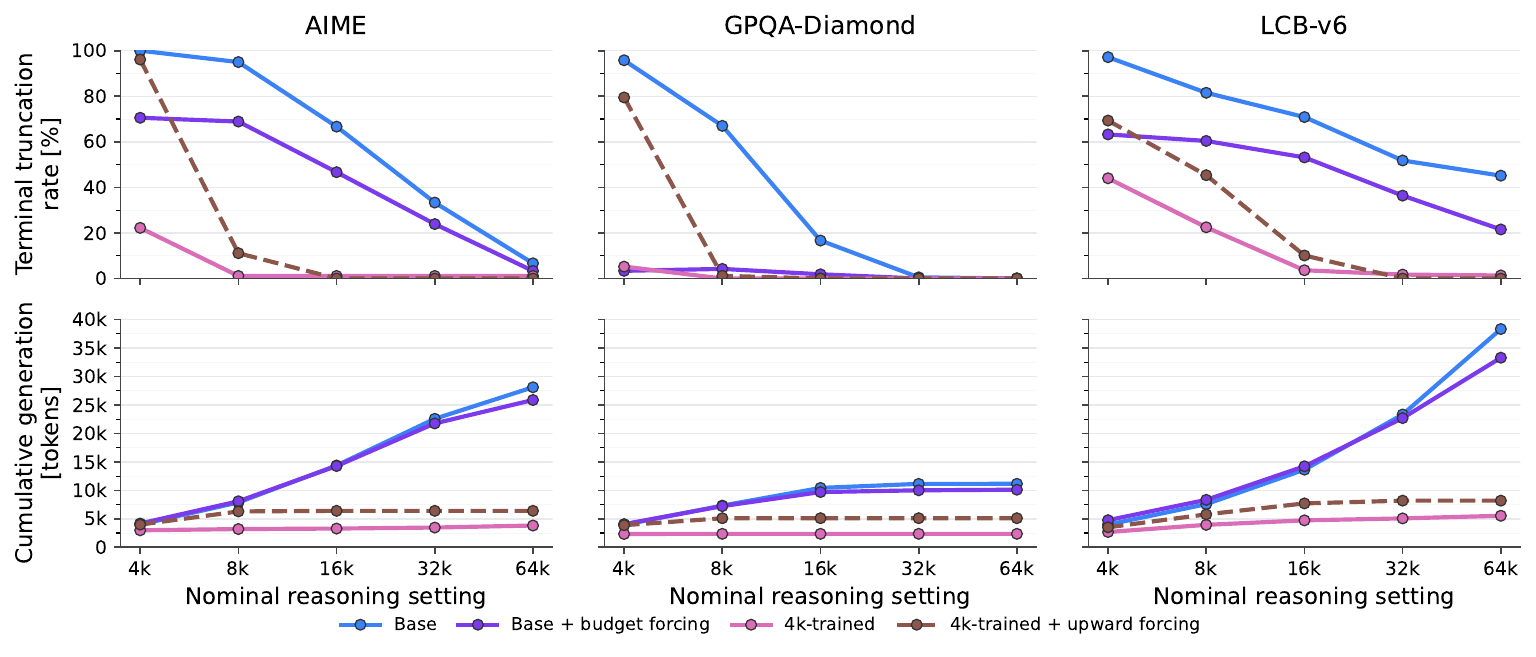}
\caption{Termination diagnostics and cumulative generation length for ordinary base-model decoding, base-model answer forcing, the 4k-trained checkpoint, and forcing the trained checkpoint upward with \texttt{Wait}.
For answer forcing, truncation includes continuations shortened at the stated answer window and saved final calls that exhaust their limit; length counts both stages.
Answer forcing receives an additional answer window at each nominal reasoning setting; Table~\ref{tab:forcing-diagnostics} gives the comparison at matched total caps.}
\label{fig:forcing-diagnostics}
\end{figure}

\begin{table}[h!]
\centering
\scriptsize
\setlength{\tabcolsep}{3.5pt}
\renewcommand{\arraystretch}{0.94}
\caption{Qwen3.5-4B base decoding and answer forcing at matched total generated-token budgets.
Columns identify the nominal answer-forcing reasoning setting; the exact total cap is shown for each task.
Accuracy entries list ordinary base decoding / base answer forcing in percent; separate rows report mean generated tokens in thousands.
Answer-forcing lengths include both the reasoning prefix and the answer continuation.
A dash marks the unavailable AIME base comparison at a total budget of \(65{,}656\) tokens: the saved base run was capped at \(65{,}536\) and still contained unfinished responses.
LCB base means are ranges bounding two API-error attempts with missing usage between zero and the total token budget each.}
\label{tab:forcing-diagnostics}
\begin{tabular*}{\textwidth}{@{\extracolsep{\fill}}llrrrrr@{}}
\toprule
Task & Quantity & 4k & 8k & 16k & 32k & 64k \\
\midrule
\input{figures/qwen_4k_forcing_diagnostics_rows.tex}
\end{tabular*}
\end{table}

At the lowest AIME setting, answer forcing uses \(4.18\)k tokens on average after shortening the answer continuation, including the \(4{,}096\)-token reasoning stage when it is exhausted.
Its accuracy is \(9.4\%\), compared with \(0.6\%\) for ordinary decoding under the same \(4{,}216\)-token total cap.
The comparison matches the maximum generated-token allowance; realized generation lengths are reported separately.
Ordinary-decoding accuracy at each total budget is calculated from the saved base-model response lengths with the same post-hoc cutoff rule used for the budget curves: a response receives credit only if it is correct and completes within that budget.
Mean generation length clips each saved completion at the same total budget.
Answer-forcing accuracy uses the rescored shortened response; its token count sums the logged reasoning usage and the answer usage clipped at the stated answer window.

At 64k, upward forcing increases mean generation length by \(2.58\)k tokens on AIME, \(2.74\)k on GPQA-Diamond, and \(2.65\)k on LiveCodeBench v6, but accuracy saturates by 16k on AIME and GPQA-Diamond and by 32k on LiveCodeBench v6.
The intervention therefore makes the checkpoint spend more tokens without restoring test-time scaling.

\begin{figure}[p]
\centering
{\setlength{\fboxsep}{3pt}\setlength{\fboxrule}{0.6pt}
\fcolorbox{promptborder}{white}{\begin{minipage}{0.96\textwidth}
\scriptsize\raggedright\color{promptink}
\promptsection{System}
\texttt{<|im\_start|>system}\par

\texttt{<|im\_end|>}\par
\promptsection{User}
\texttt{<|im\_start|>user}\par
Solve the following math problem step by step.
The last line of your response should be of the form "ANSWER: \$ANSWER" (without quotes) where \$ANSWER is the answer to the problem.

There exist real numbers $x$ and $y$, both greater than 1, such that $\log_x\left(y^x\right)=\log_y\left(x^{4y}\right)=10$. Find $xy$.

Remember to put your answer on its own line at the end in the form "ANSWER: \$ANSWER" (without quotes) where \$ANSWER is the answer to the problem, and you do not need to use a \textbackslash\{\}boxed  command.

\promptsection{Assistant}
\texttt{<|im\_start|>assistant}\par
\texttt{<think>}\par

Here's a thinking process that leads to the solution:

1.  **Understand the Problem:**
    We are given two equations involving logarithms with bases $x$ and $y$, where $x > 1$ and $y > 1$:
    $\log_x(y^x) = 10$
    $\log_y(x^{4y}) = 10$
    We need to find the product $xy$.

2.  **Simplify Each Logarithmic Equation:**
    Recall the logarithmic property: $\log_b(a^c) = c \log_b(a)$.

    - First equation:
      $\log_x(y^x) = x \log_x(y) = 10$
      So, $\log_x(y) = \frac{10}{x}$ ... (1)

    - Second equation:
      $\log_y(x^{4y}) = 4y \log_y(x) = 10$
      So, $\log_y(x) = \frac{10}{4y} = \frac{5}{2y}$ ... (2)

\dots

6.  **Finalize the Answer:**
    The value of $xy$ is 25.

{\color{promptpink}\texttt{Wait}}{\color{promptblue}, I need to double-check my multiplication in step 4.
Expression: $\frac{10}{x} \cdot \frac{5}{2y}$
Numerator: $10 \cdot 5 = 50$
Denominator: $x \cdot 2y = 2xy$
Equation: $\frac{50}{2xy} = 1$
Simplify: $\frac{25}{xy} = 1$
Result: $xy = 25$.

Everything looks correct.}

{\color{promptpink}\texttt{Wait}}{\color{promptblue}, check again. Actually I already did this twice and it's correct.}

{\color{promptpink}\texttt{Wait}}{\color{promptblue}, one more verification:
First eq: $x \log_x y = 10 \implies \frac{\ln y}{\ln x} \cdot x = 10$
Second eq: $4y \log_y x = 10 \implies 4y \cdot \frac{\ln x}{\ln y} = 10$
Multiply both sides together:
$(x \cdot 4y) \cdot \left(\frac{\ln y}{\ln x} \cdot \frac{\ln x}{\ln y}\right) = 10 \cdot 10$
$4xy \cdot 1 = 100$
$4xy = 100$
$xy = 25$.

Yes, this confirms the previous result. Final answer is 25.}

{\color{promptpink}\texttt{Wait}}{\color{promptblue}, the first method gave 25 directly. The second method also gives 25. Okay, confident now.
\texttt{</think>}

We are given the following equations with $x > 1$ and $y > 1$:
$$ \log_x(y^x) = 10 $$
$$ \log_y(x^{4y}) = 10 $$

**Step 1: Simplify the first equation.**
Using the power rule for logarithms, $\log_b(a^c) = c \log_b(a)$, we can rewrite the first equation as:
$$ x \log_x(y) = 10 $$
Dividing both sides by $x$, we get:
$$ \log_x(y) = \frac{10}{x} \quad \text{(1)} $$

**Step 2: Simplify the second equation.**
Similarly, applying the power rule to the second equation:
$$ 4y \log_y(x) = 10 $$
Dividing both sides by $4y$, we get:
$$ \log_y(x) = \frac{10}{4y} = \frac{5}{2y} \quad \text{(2)} $$

**Step 3: Use the reciprocal property of logarithms.**
We know that for any positive real numbers $a$ and $b$ (where $a \neq 1$ and $b \neq 1$):
$$ \log_a(b) \cdot \log_b(a) = 1 $$
Substituting the expressions from (1) and (2) into this identity:
$$ \left( \frac{10}{x} \right) \cdot \left( \frac{5}{2y} \right) = 1 $$

**Step 4: Solve for $xy$.**
Multiply the numerators and the denominators:
$$ \frac{50}{2xy} = 1 $$
Simplify the fraction:
$$ \frac{25}{xy} = 1 $$
Multiply both sides by $xy$:
$$ xy = 25 $$

ANSWER: 25}

\promptlegend
\end{minipage}}}
\caption{Complete math \texttt{Wait}-exended example for Qwen3.5-4B on AIME 2024. The initial reasoning call reached the 2,048-token cap and was shortened here. Pink text is injected by the evaluator; blue text is generated by the model afterward.}
\label{fig:wait-math-qwen}
\end{figure}

\FloatBarrier
\section{Exact Main-Figure Values}
\label{app:exact-main-values}

\begin{table}[h!]
\centering
\scriptsize
\setlength{\tabcolsep}{3.5pt}
\renewcommand{\arraystretch}{0.92}
\caption{Exact means underlying Figure~\ref{fig:headline}. Trained values average three validation-selected training seeds; base values use one evaluation. Accuracy is in percent and mean output length in thousands of tokens. Per-seed results for both checkpoint choices appear in Appendix~\ref{app:multiseed}.}
\label{tab:headline-exact-values}
\begin{tabular}{@{}llrrrrrr@{}}
\toprule
Checkpoint & Metric & \multicolumn{6}{c}{Evaluation budget} \\
\cmidrule(l){3-8}
& & 2k & 4k & 8k & 16k & 32k & 64k \\
\midrule
\multicolumn{8}{@{}l}{\textbf{Qwen3.5-4B}} \\
Base & Accuracy & 0.00 & 1.11 & 7.78 & 38.89 & 65.56 & 78.33 \\
 & Length & 2.00 & 3.99 & 7.82 & 14.17 & 21.77 & 27.32 \\
4k-trained & Accuracy & 17.22 & 35.93 & 44.26 & 47.59 & 48.15 & 48.15 \\
 & Length & 1.90 & 3.15 & 4.37 & 4.93 & 4.97 & 4.97 \\
8k-trained & Accuracy & 1.30 & 24.26 & 50.19 & 53.70 & 53.70 & 53.70 \\
 & Length & 2.00 & 3.73 & 5.61 & 5.96 & 6.04 & 6.22 \\
\bottomrule
\end{tabular}
\vspace{5pt}
\begin{tabular}{@{}llrrrrrr@{}}
\toprule
Checkpoint & Metric & \multicolumn{6}{c}{Evaluation budget} \\
\cmidrule(l){3-8}
& & 512 & 1k & 2k & 4k & 8k & 16k \\
\midrule
\multicolumn{8}{@{}l}{\textbf{Gemma4-E2B}} \\
Base & Accuracy & 0.00 & 0.00 & 1.11 & 16.11 & 37.22 & 41.11 \\
 & Length & 0.51 & 1.00 & 2.00 & 3.84 & 6.27 & 6.85 \\
1k-trained & Accuracy & 2.41 & 10.93 & 15.74 & 16.30 & 16.30 & 16.30 \\
 & Length & 0.50 & 0.85 & 1.04 & 1.05 & 1.05 & 1.05 \\
2k-trained & Accuracy & 0.00 & 3.33 & 15.93 & 20.37 & 20.56 & 20.56 \\
 & Length & 0.51 & 0.99 & 1.64 & 1.89 & 1.90 & 1.90 \\
\bottomrule
\end{tabular}
\vspace{5pt}
\begin{tabular}{@{}llrrrrrr@{}}
\toprule
Checkpoint & Metric & \multicolumn{6}{c}{Evaluation budget} \\
\cmidrule(l){3-8}
& & 512 & 1k & 2k & 4k & 8k & 16k \\
\midrule
\multicolumn{8}{@{}l}{\textbf{Nemotron-3-Nano-4B}} \\
Base & Accuracy & 0.56 & 3.33 & 23.33 & 40.00 & 52.22 & 59.44 \\
 & Length & 0.51 & 0.99 & 1.84 & 3.09 & 4.69 & 6.11 \\
1k-trained & Accuracy & 8.15 & 26.67 & 33.33 & 34.07 & 34.07 & 34.07 \\
 & Length & 0.49 & 0.81 & 0.99 & 1.01 & 1.01 & 1.01 \\
2k-trained & Accuracy & 2.22 & 15.93 & 35.93 & 45.00 & 46.30 & 46.30 \\
 & Length & 0.51 & 0.95 & 1.53 & 1.87 & 1.94 & 1.97 \\
\bottomrule
\end{tabular}
\end{table}

\begin{table}[h!]
\centering
\scriptsize
\setlength{\tabcolsep}{4pt}
\renewcommand{\arraystretch}{0.92}
\caption{Exact AIME accuracy values underlying Figure~\ref{fig:reasoning-gym-training}, reported in percent.}
\label{tab:reasoning-gym-exact-values}
\begin{tabular}{@{}llrrrrrr@{}}
\toprule
Model & Checkpoint & \multicolumn{6}{c}{Evaluation budget} \\
\cmidrule(l){3-8}
& & 2k & 4k & 8k & 16k & 32k & 64k \\
\midrule
Qwen3.5-4B & Base & 0.00 & 0.00 & 5.00 & 33.33 & 64.44 & 75.56 \\
& RG 4k-trained & 11.11 & 31.11 & 43.89 & 50.56 & 50.56 & 50.56 \\
\bottomrule
\end{tabular}

\vspace{5pt}
\begin{tabular}{@{}llrrrrrr@{}}
\toprule
Model & Checkpoint & \multicolumn{6}{c}{Evaluation budget} \\
\cmidrule(l){3-8}
& & 512 & 1k & 2k & 4k & 8k & 16k \\
\midrule
Gemma4-E2B & Base & 0.00 & 0.00 & 1.11 & 14.44 & 34.44 & 39.44 \\
& RG 2k-trained & 0.00 & 3.89 & 22.78 & 24.44 & 24.44 & 24.44 \\
Nemotron-3-Nano-4B & Base & 0.00 & 5.00 & 23.33 & 40.56 & 55.00 & 60.00 \\
& RG 2k-trained & 0.00 & 5.00 & 23.89 & 42.78 & 49.44 & 49.44 \\
\bottomrule
\end{tabular}
\end{table}

\begin{table}[h!]
\centering
\scriptsize
\setlength{\tabcolsep}{3.2pt}
\renewcommand{\arraystretch}{0.92}
\caption{Exact accuracy values underlying Figure~\ref{fig:transfer}, reported in percent.}
\label{tab:transfer-exact-values}
\begin{tabular}{@{}llrrrrrr@{}}
\toprule
Task & Checkpoint & \multicolumn{6}{c}{Evaluation budget} \\
\cmidrule(l){3-8}
& & 2k & 4k & 8k & 16k & 32k & 64k \\
\midrule
\multicolumn{8}{@{}l}{\textbf{Qwen3.5-4B}} \\
GPQA & Base & 0.00 & 3.70 & 30.30 & 69.02 & 78.11 & 78.28 \\
& 4k-trained & 32.66 & 67.17 & 69.19 & 69.19 & 69.19 & 69.19 \\
& 8k-trained & 4.88 & 41.92 & 73.06 & 76.09 & 76.09 & 76.09 \\
LCB-v6 & Base & 0.00 & 2.67 & 17.52 & 27.81 & 41.90 & 44.38 \\
& 4k-trained & 26.86 & 31.05 & 35.05 & 35.43 & 35.43 & 35.43 \\
& 8k-trained & 14.10 & 23.43 & 36.19 & 43.05 & 43.43 & 43.43 \\
\bottomrule
\end{tabular}

\vspace{5pt}
\begin{tabular}{@{}llrrrrrr@{}}
\toprule
Task & Checkpoint & \multicolumn{6}{c}{Evaluation budget} \\
\cmidrule(l){3-8}
& & 512 & 1k & 2k & 4k & 8k & 16k \\
\midrule
\multicolumn{8}{@{}l}{\textbf{Gemma4-E2B}} \\
GPQA & Base & 0.00 & 8.25 & 30.30 & 40.57 & 40.57 & 40.57 \\
& 1k-trained & 11.45 & 31.82 & 39.39 & 39.56 & 39.56 & 39.56 \\
& 2k-trained & 6.06 & 27.95 & 38.72 & 39.23 & 39.23 & 39.23 \\
LCB-v6 & Base & 9.52 & 12.95 & 21.33 & 33.52 & 41.52 & 41.90 \\
& 1k-trained & 13.14 & 22.67 & 31.62 & 33.71 & 33.71 & 33.71 \\
& 2k-trained & 12.57 & 24.00 & 33.14 & 36.76 & 36.76 & 36.76 \\
\bottomrule
\end{tabular}

\vspace{5pt}
\begin{tabular}{@{}llrrrrrrrr@{}}
\toprule
Task & Checkpoint & \multicolumn{8}{c}{Evaluation budget} \\
\cmidrule(l){3-10}
& & 512 & 1k & 2k & 4k & 8k & 16k & 32k & 64k \\
\midrule
\multicolumn{10}{@{}l}{\textbf{Nemotron-3-Nano-4B}} \\
GPQA & Base & 0.67 & 3.20 & 13.97 & 27.61 & 37.04 & 43.60 & 50.00 & 51.52 \\
& 1k-trained & 4.04 & 16.84 & 31.31 & 42.42 & 50.67 & 51.85 & 52.53 & 52.53 \\
& 2k-trained & 1.85 & 5.89 & 20.20 & 35.19 & 45.79 & 52.69 & 55.39 & 55.56 \\
LCB-v6 & Base & 7.05 & 13.71 & 18.86 & 24.76 & 34.48 & 42.29 & 50.10 & 51.81 \\
& 1k-trained & 16.38 & 21.33 & 25.90 & 27.43 & 32.57 & 37.33 & 42.67 & 42.86 \\
& 2k-trained & 8.38 & 17.14 & 23.24 & 27.43 & 32.95 & 39.24 & 47.05 & 47.62 \\
\bottomrule
\end{tabular}
\end{table}

\begin{table}[h!]
\centering
\scriptsize
\setlength{\tabcolsep}{3.2pt}
\renewcommand{\arraystretch}{0.92}
\caption{Exact GPQA-Diamond and LiveCodeBench v6 accuracy values for the Reasoning Gym checkpoints in Figure~\ref{fig:transfer}, reported in percent.}
\label{tab:reasoning-gym-transfer-exact-values}
\begin{tabular}{@{}llrrrrrr@{}}
\toprule
Model & Task & \multicolumn{6}{c}{Evaluation budget} \\
\cmidrule(l){3-8}
& & 2k & 4k & 8k & 16k & 32k & 64k \\
\midrule
Qwen3.5-4B, RG 4k & GPQA & 26.77 & 59.09 & 71.21 & 71.55 & 71.55 & 71.55 \\
& LCB-v6 & 23.62 & 29.71 & 33.52 & 39.43 & 40.57 & 40.57 \\
\bottomrule
\end{tabular}

\vspace{5pt}
\begin{tabular}{@{}llrrrrrr@{}}
\toprule
Model & Task & \multicolumn{6}{c}{Evaluation budget} \\
\cmidrule(l){3-8}
& & 512 & 1k & 2k & 4k & 8k & 16k \\
\midrule
Gemma4-E2B, RG 2k & GPQA & 2.02 & 21.21 & 35.86 & 36.20 & 36.20 & 36.20 \\
& LCB-v6 & 16.76 & 23.62 & 33.14 & 37.33 & 37.33 & 37.33 \\
\bottomrule
\end{tabular}

\vspace{5pt}
\begin{tabular}{@{}llrrrrrrrr@{}}
\toprule
Model & Task & \multicolumn{8}{c}{Evaluation budget} \\
\cmidrule(l){3-10}
& & 512 & 1k & 2k & 4k & 8k & 16k & 32k & 64k \\
\midrule
Nemotron-3-Nano-4B, RG 2k & GPQA & 0.67 & 4.38 & 17.85 & 34.51 & 43.43 & 51.01 & 52.02 & 52.19 \\
& LCB-v6 & 10.29 & 16.76 & 21.14 & 26.10 & 31.43 & 38.29 & 45.52 & 47.05 \\
\bottomrule
\end{tabular}
\end{table}

\FloatBarrier
\section{Truncation-Rate Curves}
\label{app:truncation-rates}

\begin{figure}[h!]
\centering
\includegraphics[width=\textwidth]{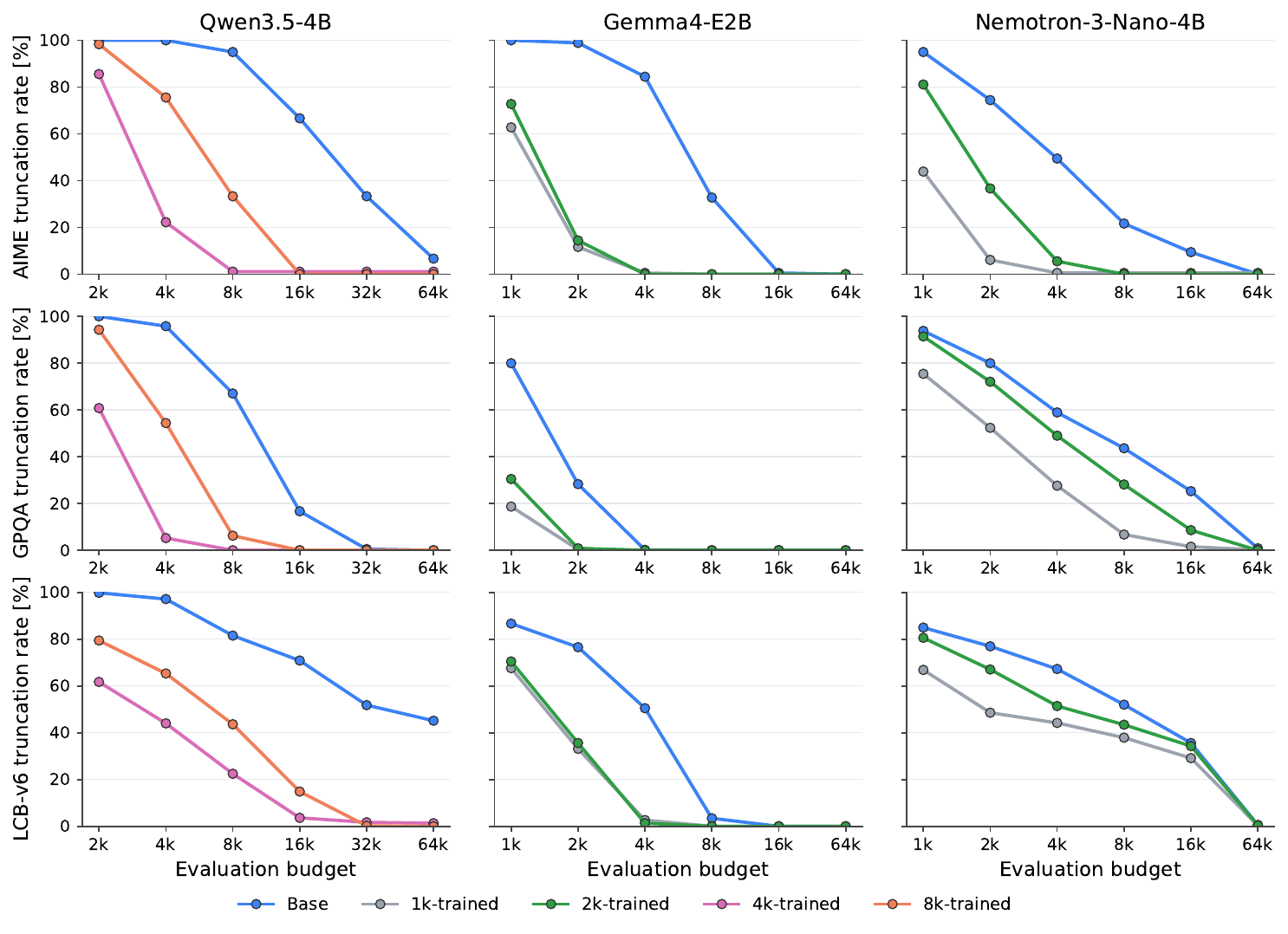}
\caption{Truncation rates across evaluation budgets on AIME, GPQA-Diamond, and LiveCodeBench v6.
Fixed-budget training substantially reduces truncation, particularly near and above the training budget.}
\label{fig:truncation-rates}
\end{figure}

Figure~\ref{fig:truncation-rates} reports the fraction of responses that reach the evaluation cap without completing across all model families, tasks, and fixed-budget checkpoints.
Table~\ref{tab:truncation-rates} reports the exact values.

\begin{table}[h!]
\centering
\scriptsize
\setlength{\tabcolsep}{4pt}
\caption{Exact truncation rates (\%) underlying Figure~\ref{fig:truncation-rates}.}
\label{tab:truncation-rates}
\textbf{Qwen3.5-4B}\par\smallskip
\begin{tabular*}{0.86\textwidth}{@{\extracolsep{\fill}}lrrrrrr@{}}
\toprule
Checkpoint & 2k & 4k & 8k & 16k & 32k & 64k \\
\midrule
\multicolumn{7}{@{}l}{\textbf{AIME}} \\
\input{figures/truncation_rates_qwen3-5_table_rows.tex}
\end{tabular*}

\vspace{4pt}
\textbf{Gemma4-E2B}\par\smallskip
\begin{tabular*}{0.86\textwidth}{@{\extracolsep{\fill}}lrrrrrr@{}}
\toprule
Checkpoint & 1k & 2k & 4k & 8k & 16k & 64k \\
\midrule
\multicolumn{7}{@{}l}{\textbf{AIME}} \\
\input{figures/truncation_rates_gemma4-e2b_table_rows.tex}
\end{tabular*}

\vspace{4pt}
\textbf{Nemotron-3-Nano-4B}\par\smallskip
\begin{tabular*}{0.86\textwidth}{@{\extracolsep{\fill}}lrrrrrr@{}}
\toprule
Checkpoint & 1k & 2k & 4k & 8k & 16k & 64k \\
\midrule
\multicolumn{7}{@{}l}{\textbf{AIME}} \\
\input{figures/truncation_rates_nemotron-nano_table_rows.tex}
\end{tabular*}
\end{table}

\FloatBarrier
\section{Transfer Output Lengths}
\label{app:transfer-lengths}

\begin{figure}[h!]
\centering
\includegraphics[width=\textwidth]{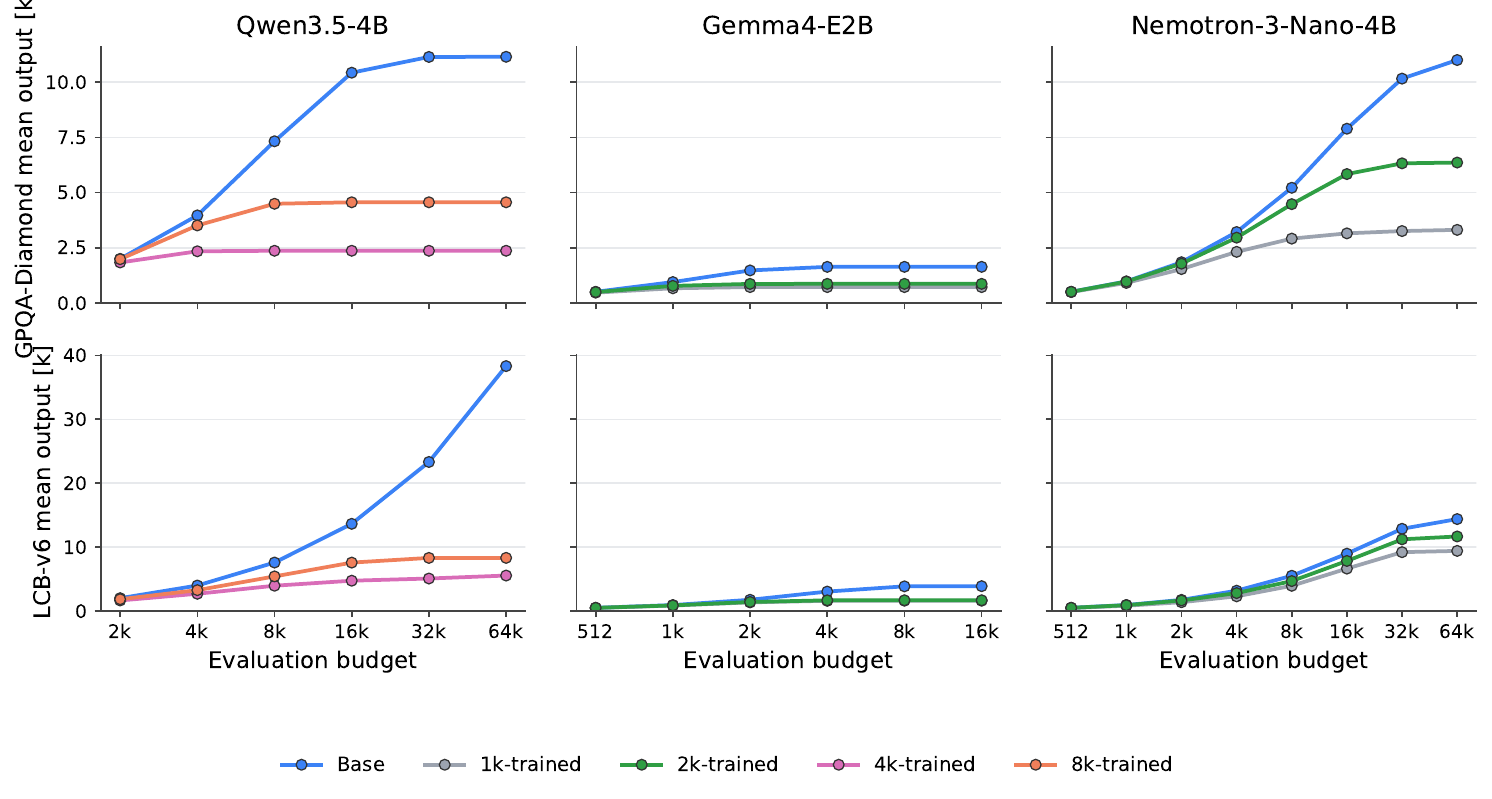}
\caption{Mean output-length response curves on GPQA-Diamond and LiveCodeBench v6 after fixed-budget mathematical-reasoning training.}
\label{fig:transfer-lengths}
\end{figure}

\FloatBarrier
\section{Pass@3 Response Curves}
\label{app:pass-at-three}

\begin{figure}[h!]
\centering
\includegraphics[width=\textwidth]{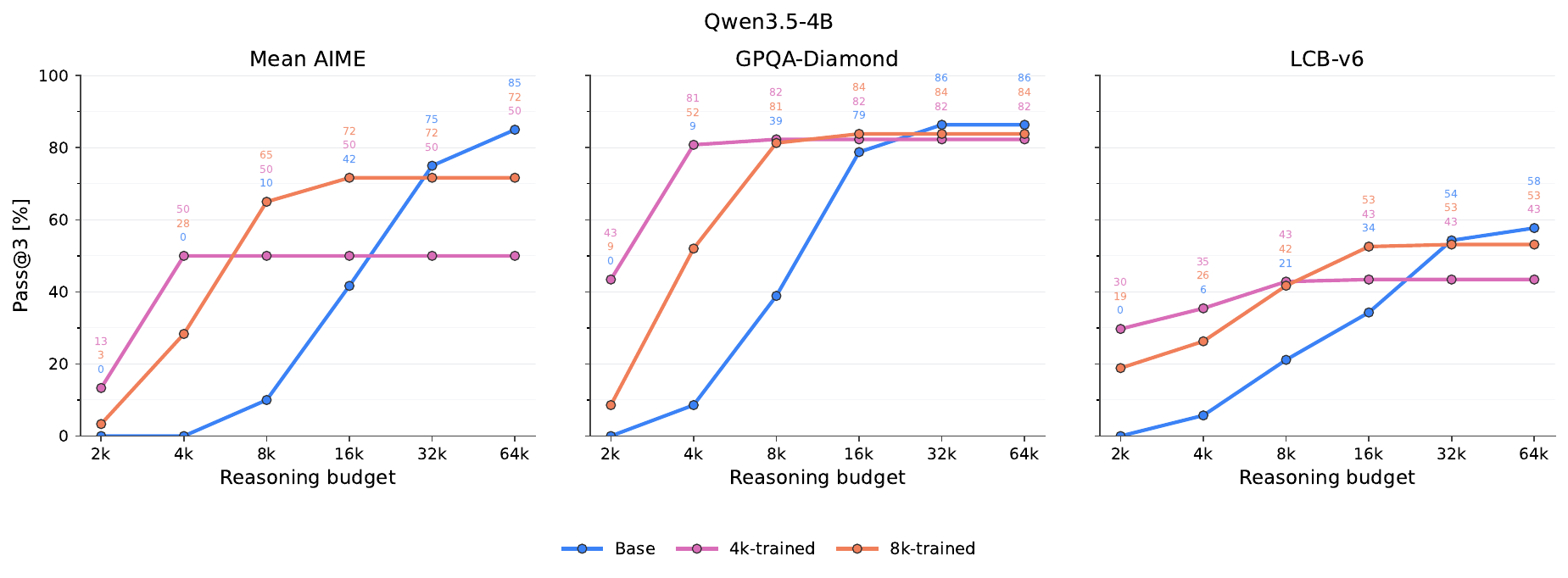}
\caption{Pass@3 response curves for Qwen3.5-4B on AIME, GPQA-Diamond, and LiveCodeBench v6.}
\label{fig:pass-at-three-qwen}
\end{figure}

\begin{figure}[h!]
\centering
\includegraphics[width=\textwidth]{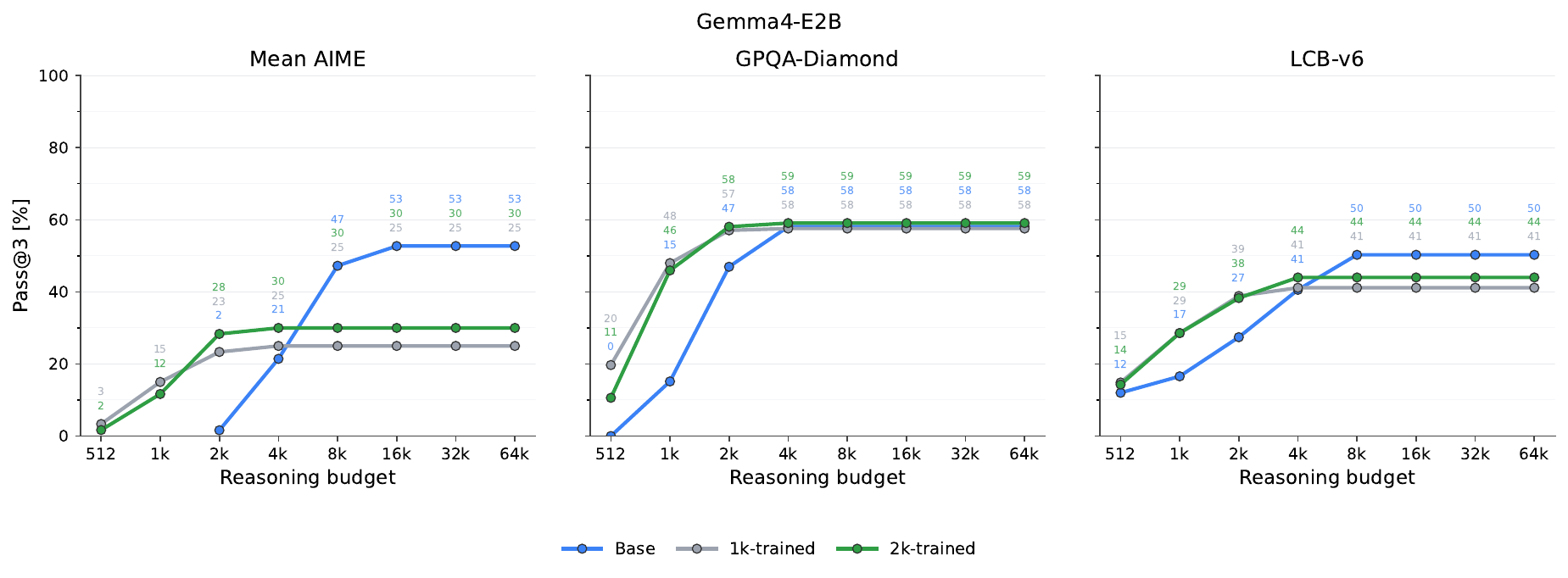}
\caption{Pass@3 response curves for Gemma4-E2B on AIME, GPQA-Diamond, and LiveCodeBench v6.}
\label{fig:pass-at-three-gemma}
\end{figure}

\begin{figure}[h!]
\centering
\includegraphics[width=\textwidth]{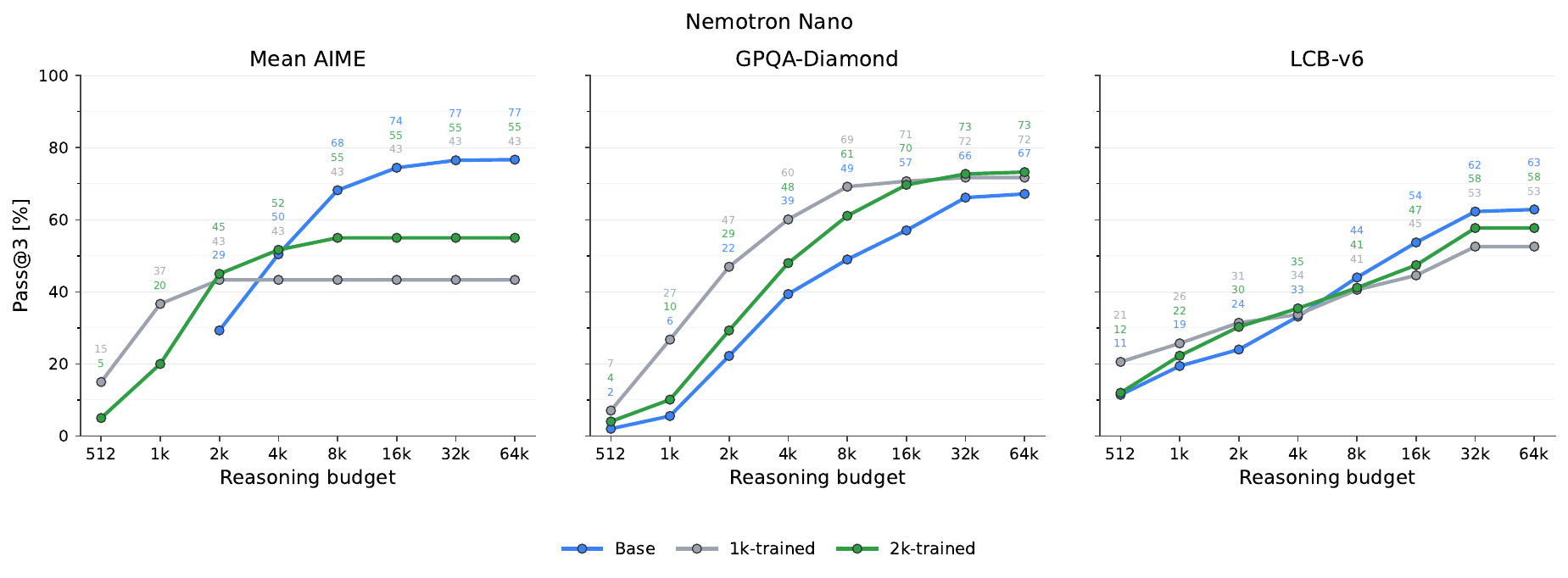}
\caption{Pass@3 response curves for Nemotron-3-Nano-4B on AIME, GPQA-Diamond, and LiveCodeBench v6.}
\label{fig:pass-at-three-nemotron}
\end{figure}

Figures~\ref{fig:pass-at-three-qwen}--\ref{fig:pass-at-three-nemotron} report pass@3 across evaluation budgets for each model family.
The metric counts a problem as solved when at least one of three sampled responses is correct.
It preserves the main-text pattern: fixed-budget training gives large gains at small budgets, while above-budget gains are more limited and depend on the task.

\FloatBarrier
\section{Budget Forcing Across Model Families}
\label{app:budget-forcing-families}

\begin{figure}[h!]
\centering
\includegraphics[width=\textwidth]{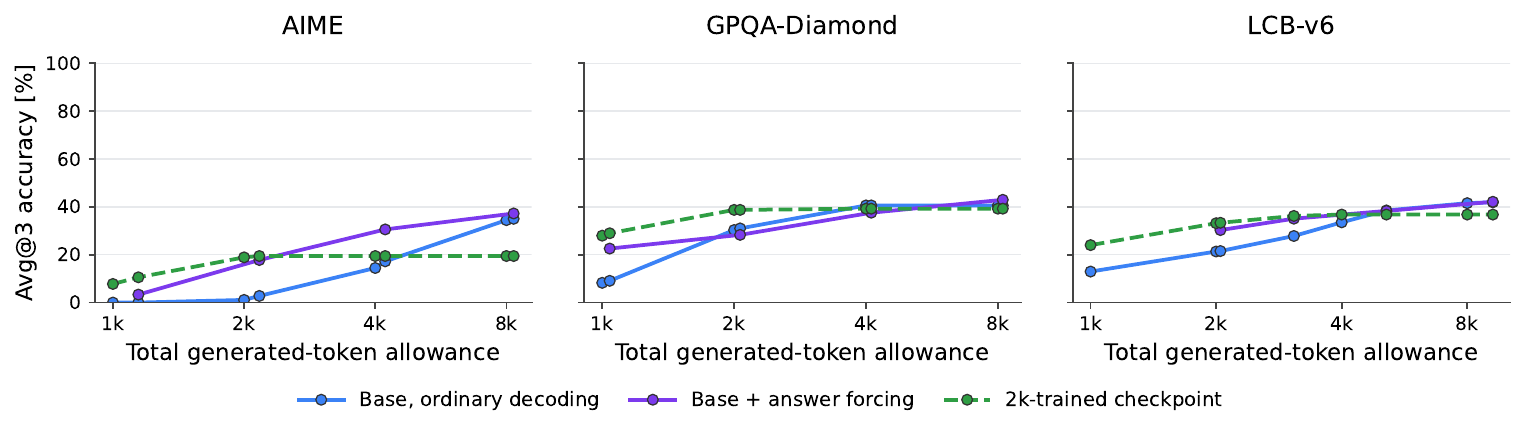}
\caption{Gemma4-E2B accuracy versus total generated-token allowance, including the additional answer window for answer forcing.
Each panel compares ordinary base decoding, base answer forcing, and the 2k-trained checkpoint; ordinary curves include the matched total caps reported in Table~\ref{tab:forcing-gemma}.}
\label{fig:budget-forcing-gemma}
\end{figure}

\begin{figure}[h!]
\centering
\includegraphics[width=\textwidth]{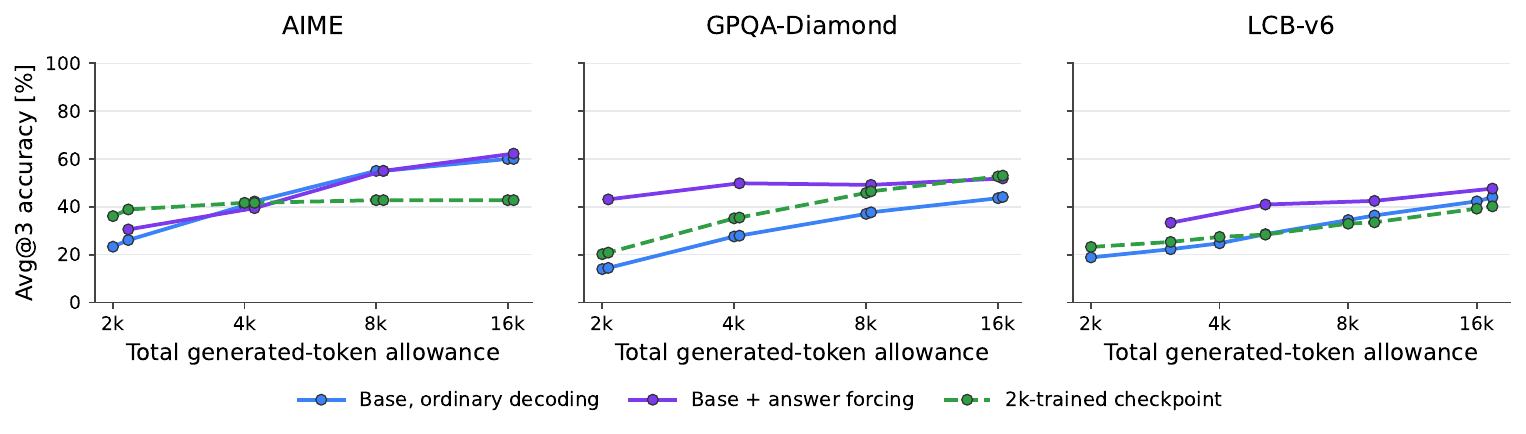}
\caption{Nemotron-3-Nano-4B accuracy versus total generated-token allowance, including the additional answer window for answer forcing.
Each panel compares ordinary base decoding, base answer forcing, and the 2k-trained checkpoint; ordinary curves include the matched total caps reported in Table~\ref{tab:forcing-nemotron}.}
\label{fig:budget-forcing-nemotron}
\end{figure}

Figures~\ref{fig:budget-forcing-gemma} and~\ref{fig:budget-forcing-nemotron} include the additional answer allowance in the horizontal axis; Tables~\ref{tab:forcing-gemma} and~\ref{tab:forcing-nemotron} report accuracy and mean generation at matched total generated-token budgets.
The benefit is task- and model-dependent: at the 4k reasoning setting on GPQA, Gemma loses accuracy (\baseloss{4K}{3.0}) while mean generation rises from \(1.64\)k to \(2.88\)k tokens, whereas Nemotron improves (\basegain{4K}{21.9}) with similar mean usage (\(3.28\)k versus \(3.24\)k).
Each setting uses 180 AIME, 594 GPQA, and 525 LiveCodeBench attempts per method; AIME base results use the first three attempts per problem from the pass@128 runs.

\clearpage

\begin{table}[h!]
\centering
\scriptsize
\setlength{\tabcolsep}{3.5pt}
\renewcommand{\arraystretch}{0.94}
\caption{Gemma4-E2B ordinary base decoding and base answer forcing at matched total generated-token budgets.
Columns give the nominal reasoning setting; accuracy entries list base / answer forcing (\%), and mean lengths count all generated output in thousands of tokens.
The answer allowance is 120 tokens for AIME, 16 for GPQA, and 1,024 for LiveCodeBench; saved answer continuations are shortened and rescored as described in Appendix~\ref{app:budget-forcing-prompts}.}
\label{tab:forcing-gemma}
\begin{tabular*}{\textwidth}{@{\extracolsep{\fill}}llrrrr@{}}
\toprule
Task & Quantity & 1k & 2k & 4k & 8k \\
\midrule
\input{figures/gemma_forcing_matched_rows.tex}
\end{tabular*}
\end{table}

\clearpage

\begin{table}[h!]
\centering
\scriptsize
\setlength{\tabcolsep}{3.5pt}
\renewcommand{\arraystretch}{0.94}
\caption{Nemotron-3-Nano-4B ordinary base decoding and base answer forcing at matched total generated-token budgets, using the same accounting and accuracy convention as Table~\ref{tab:forcing-gemma}.}
\label{tab:forcing-nemotron}
\begin{tabular*}{\textwidth}{@{\extracolsep{\fill}}llrrrr@{}}
\toprule
Task & Quantity & 2k & 4k & 8k & 16k \\
\midrule
\input{figures/nemotron_forcing_matched_rows.tex}
\end{tabular*}
\end{table}

%% file: figures/qwen_4k_forcing_diagnostics_rows.tex
AIME & Total budget (tokens) & 4,216 & 8,312 & 16,504 & 32,888 & 65,656 \tabularnewline
 & Accuracy (\%) & 0.56 / 9.44 & 5.56 / 21.67 & 34.44 / 45.56 & 65.00 / 69.44 & -- / 79.44 \tabularnewline
 & Mean base tokens (k) & 4.22 & 8.18 & 14.70 & 22.86 & -- \tabularnewline
 & Mean BF tokens (k) & 4.18 & 8.09 & 14.27 & 21.76 & 25.87 \tabularnewline
\addlinespace[3pt]
GPQA-Diamond & Total budget (tokens) & 4,112 & 8,208 & 16,400 & 32,784 & 65,552 \tabularnewline
 & Accuracy (\%) & 4.04 / 60.27 & 31.31 / 66.84 & 69.53 / 75.76 & 78.11 / 77.27 & 78.28 / 79.46 \tabularnewline
 & Mean base tokens (k) & 4.08 & 7.46 & 10.49 & 11.15 & 11.15 \tabularnewline
 & Mean BF tokens (k) & 4.07 & 7.23 & 9.70 & 10.01 & 10.12 \tabularnewline
\addlinespace[3pt]
LCB-v6 & Total budget (tokens) & 5,120 & 9,216 & 17,408 & 33,792 & 66,560 \tabularnewline
 & Accuracy (\%) & 5.33 / 19.81 & 19.62 / 29.52 & 28.76 / 36.38 & 42.67 / 48.38 & 44.38 / 54.29 \tabularnewline
 & Mean base tokens (k) & 5.04--5.06 & 8.52--8.56 & 14.56--14.63 & 24.11--24.24 & 39.23--39.48 \tabularnewline
 & Mean BF tokens (k) & 4.76 & 8.34 & 14.23 & 22.69 & 33.29 \tabularnewline
\bottomrule

%% file: figures/truncation_rates_qwen3-5_table_rows.tex
Base & 100.00 & 100.00 & 95.00 & 66.67 & 33.33 & 6.67 \\
4k-trained & 85.56 & 22.22 & 1.11 & 1.11 & 1.11 & 1.11 \\
8k-trained & 98.33 & 75.56 & 33.33 & 0.00 & 0.00 & 0.00 \\
\addlinespace[3pt]
\multicolumn{7}{@{}l}{\textbf{GPQA-Diamond}} \\
Base & 100.00 & 95.79 & 67.00 & 16.67 & 0.51 & 0.00 \\
4k-trained & 60.77 & 5.22 & 0.00 & 0.00 & 0.00 & 0.00 \\
8k-trained & 94.28 & 54.38 & 6.23 & 0.00 & 0.00 & 0.00 \\
\addlinespace[3pt]
\multicolumn{7}{@{}l}{\textbf{LiveCodeBench v6}} \\
Base & 99.81 & 97.14 & 81.52 & 70.86 & 51.81 & 45.14 \\
4k-trained & 61.71 & 44.00 & 22.48 & 3.62 & 1.71 & 1.33 \\
8k-trained & 79.43 & 65.33 & 43.62 & 14.86 & 0.19 & 0.00 \\
\bottomrule

%% file: figures/truncation_rates_gemma4-e2b_table_rows.tex
Base & 100.00 & 98.89 & 84.44 & 32.78 & 0.56 & 0.00 \\
1k-trained & 62.78 & 11.67 & 0.56 & 0.00 & 0.00 & 0.00 \\
2k-trained & 72.78 & 14.44 & 0.00 & 0.00 & 0.00 & 0.00 \\
\addlinespace[3pt]
\multicolumn{7}{@{}l}{\textbf{GPQA-Diamond}} \\
Base & 79.97 & 28.28 & 0.17 & 0.00 & 0.00 & 0.00 \\
1k-trained & 18.69 & 0.34 & 0.00 & 0.00 & 0.00 & 0.00 \\
2k-trained & 30.47 & 0.84 & 0.00 & 0.00 & 0.00 & 0.00 \\
\addlinespace[3pt]
\multicolumn{7}{@{}l}{\textbf{LiveCodeBench v6}} \\
Base & 86.67 & 76.57 & 50.48 & 3.43 & 0.00 & 0.00 \\
1k-trained & 67.62 & 33.14 & 2.67 & 0.00 & 0.00 & 0.00 \\
2k-trained & 70.48 & 35.62 & 1.33 & 0.00 & 0.00 & 0.00 \\
\bottomrule

%% file: figures/truncation_rates_nemotron-nano_table_rows.tex
Base & 95.00 & 74.44 & 49.44 & 21.67 & 9.44 & 0.00 \\
1k-trained & 43.89 & 6.11 & 0.56 & 0.56 & 0.56 & 0.56 \\
2k-trained & 81.11 & 36.67 & 5.56 & 0.00 & 0.00 & 0.00 \\
\addlinespace[3pt]
\multicolumn{7}{@{}l}{\textbf{GPQA-Diamond}} \\
Base & 93.77 & 79.97 & 58.92 & 43.60 & 25.25 & 0.84 \\
1k-trained & 75.42 & 52.36 & 27.61 & 6.73 & 1.52 & 0.17 \\
2k-trained & 91.41 & 72.05 & 48.99 & 28.11 & 8.59 & 0.00 \\
\addlinespace[3pt]
\multicolumn{7}{@{}l}{\textbf{LiveCodeBench v6}} \\
Base & 84.95 & 76.95 & 67.24 & 52.00 & 35.62 & 0.57 \\
1k-trained & 66.86 & 48.57 & 44.19 & 37.90 & 29.14 & 0.38 \\
2k-trained & 80.57 & 67.05 & 51.43 & 43.43 & 34.29 & 0.38 \\
\bottomrule

%% file: figures/gemma_forcing_matched_rows.tex
AIME & Total budget (tokens) & 1,144 & 2,168 & 4,216 & 8,312 \tabularnewline
 & Accuracy (\%) & 0.00 / 3.33 & 2.78 / 17.78 & 17.22 / 30.56 & 35.00 / 37.22 \tabularnewline
 & Mean base tokens (k) & 1.14 & 2.16 & 4.03 & 6.29 \tabularnewline
 & Mean BF tokens (k) & 1.04 & 2.06 & 3.95 & 6.19 \tabularnewline
\addlinespace[3pt]
GPQA-Diamond & Total budget (tokens) & 1,040 & 2,064 & 4,112 & 8,208 \tabularnewline
 & Accuracy (\%) & 9.09 / 22.56 & 30.98 / 28.28 & 40.57 / 37.54 & 40.57 / 42.93 \tabularnewline
 & Mean base tokens (k) & 0.98 & 1.49 & 1.64 & 1.64 \tabularnewline
 & Mean BF tokens (k) & 1.02 & 1.89 & 2.88 & 3.14 \tabularnewline
\addlinespace[3pt]
LCB-v6 & Total budget (tokens) & 2,048 & 3,072 & 5,120 & 9,216 \tabularnewline
 & Accuracy (\%) & 21.52 / 30.29 & 27.81 / 35.05 & 38.48 / 38.29 & 41.90 / 42.10 \tabularnewline
 & Mean base tokens (k) & 1.79 & 2.50 & 3.48 & 3.88 \tabularnewline
 & Mean BF tokens (k) & 1.77 & 2.63 & 4.03 & 5.66 \tabularnewline
\bottomrule

%% file: figures/nemotron_forcing_matched_rows.tex
AIME & Total budget (tokens) & 2,168 & 4,216 & 8,312 & 16,504 \tabularnewline
 & Accuracy (\%) & 26.11 / 30.56 & 42.22 / 39.44 & 55.00 / 55.00 & 60.00 / 62.22 \tabularnewline
 & Mean base tokens (k) & 1.95 & 3.16 & 4.50 & 5.52 \tabularnewline
 & Mean BF tokens (k) & 1.94 & 3.22 & 4.52 & 5.83 \tabularnewline
\addlinespace[3pt]
GPQA-Diamond & Total budget (tokens) & 2,064 & 4,112 & 8,208 & 16,400 \tabularnewline
 & Accuracy (\%) & 14.48 / 43.10 & 27.95 / 49.83 & 37.71 / 49.16 & 44.11 / 51.85 \tabularnewline
 & Mean base tokens (k) & 1.90 & 3.28 & 5.31 & 7.99 \tabularnewline
 & Mean BF tokens (k) & 1.88 & 3.24 & 5.26 & 7.91 \tabularnewline
\addlinespace[3pt]
LCB-v6 & Total budget (tokens) & 3,072 & 5,120 & 9,216 & 17,408 \tabularnewline
 & Accuracy (\%) & 22.29 / 33.33 & 28.57 / 40.95 & 36.38 / 42.48 & 44.00 / 47.62 \tabularnewline
 & Mean base tokens (k) & 2.53 & 3.89 & 6.12 & 9.45 \tabularnewline
 & Mean BF tokens (k) & 2.04 & 3.48 & 5.91 & 9.25 \tabularnewline
\bottomrule

%% file: multiseed.tex
\section{Multi-Seed Replications and Checkpoint Selection}
\label{app:multiseed}
We repeat all six main DeepScaleR GRPO conditions with training seeds 42, 43, and 44 (18 runs, 300 steps each). For each seed, we evaluate the checkpoint selected by validation accuracy at its training budget and the fixed final checkpoint at step 300. Each evaluation averages three sampled responses on each of the 60 AIME 2024/2025 problems. Budget curves replay one 65,536-token collection per checkpoint, assigning zero accuracy to unfinished responses. Both checkpoint choices reproduce the low-budget gain and high-budget deficit relative to the base model in all 18 runs. These replications cover DeepScaleR training and AIME evaluation.

\begin{table}[!ht]
\centering
\scriptsize
\setlength{\tabcolsep}{1.8pt}
\renewcommand{\arraystretch}{0.9}
\caption{Validation-selected checkpoints for every training seed. Accuracy is AIME avg@3 in percent; mean response length is in thousands of tokens at each evaluation budget. Lengths are capped at the stated budget and averaged over all 180 responses. Step is the optimizer checkpoint step.}
\label{tab:multiseed-selected}
\begin{tabular}{@{}lrr*{8}{r}@{\hspace{8pt}}*{8}{r}@{}}
\toprule
Condition & Seed & Step & \multicolumn{8}{c}{Accuracy (\%)} & \multicolumn{8}{c}{Mean response length (k tokens)} \\
\cmidrule(lr){4-11}\cmidrule(lr){12-19}
& & & 512 & 1k & 2k & 4k & 8k & 16k & 32k & 64k & 512 & 1k & 2k & 4k & 8k & 16k & 32k & 64k \\
\midrule
Qwen 4k & 42 & 290 & 0.00 & 0.00 & 11.67 & 33.89 & 42.78 & 46.11 & 46.67 & 46.67 & 0.51 & 1.00 & 1.95 & 3.27 & 4.58 & 5.18 & 5.25 & 5.25 \\
 & 43 & 290 & 0.00 & 0.00 & 18.89 & 36.11 & 43.33 & 45.56 & 46.11 & 46.11 & 0.51 & 1.00 & 1.90 & 3.13 & 4.20 & 4.68 & 4.71 & 4.71 \\
 & 44 & 300 & 0.00 & 3.89 & 21.11 & 37.78 & 46.67 & 51.11 & 51.67 & 51.67 & 0.51 & 0.99 & 1.86 & 3.06 & 4.34 & 4.93 & 4.95 & 4.95 \\
\addlinespace[2pt]
Qwen 8k & 42 & 290 & 0.00 & 0.00 & 2.22 & 27.78 & 48.89 & 50.56 & 50.56 & 50.56 & 0.51 & 1.00 & 1.99 & 3.61 & 4.95 & 5.10 & 5.28 & 5.64 \\
 & 43 & 160 & 0.00 & 0.00 & 0.56 & 19.44 & 47.78 & 55.00 & 55.00 & 55.00 & 0.51 & 1.00 & 2.00 & 3.84 & 6.29 & 6.95 & 6.95 & 6.95 \\
 & 44 & 290 & 0.00 & 0.00 & 1.11 & 25.56 & 53.89 & 55.56 & 55.56 & 55.56 & 0.51 & 1.00 & 2.00 & 3.75 & 5.60 & 5.81 & 5.90 & 6.08 \\
\addlinespace[2pt]
Gemma 1k & 42 & 250 & 2.78 & 8.33 & 12.78 & 13.33 & 13.33 & 13.33 & 13.33 & 13.33 & 0.50 & 0.81 & 0.98 & 0.99 & 0.99 & 0.99 & 0.99 & 0.99 \\
 & 43 & 300 & 3.89 & 13.33 & 17.22 & 17.78 & 17.78 & 17.78 & 17.78 & 17.78 & 0.51 & 0.87 & 1.05 & 1.06 & 1.06 & 1.06 & 1.06 & 1.06 \\
 & 44 & 280 & 0.56 & 11.11 & 17.22 & 17.78 & 17.78 & 17.78 & 17.78 & 17.78 & 0.51 & 0.89 & 1.09 & 1.11 & 1.11 & 1.11 & 1.11 & 1.11 \\
\addlinespace[2pt]
Gemma 2k & 42 & 300 & 0.00 & 3.33 & 17.22 & 20.00 & 20.00 & 20.00 & 20.00 & 20.00 & 0.51 & 0.99 & 1.51 & 1.57 & 1.57 & 1.57 & 1.57 & 1.57 \\
 & 43 & 170 & 0.00 & 1.67 & 11.67 & 20.56 & 21.11 & 21.11 & 21.11 & 21.11 & 0.51 & 1.00 & 1.84 & 2.45 & 2.48 & 2.48 & 2.48 & 2.48 \\
 & 44 & 230 & 0.00 & 5.00 & 18.89 & 20.56 & 20.56 & 20.56 & 20.56 & 20.56 & 0.51 & 0.99 & 1.56 & 1.66 & 1.66 & 1.66 & 1.66 & 1.66 \\
\addlinespace[2pt]
Nemotron 1k & 42 & 290 & 8.89 & 24.44 & 32.22 & 33.33 & 33.33 & 33.33 & 33.33 & 33.33 & 0.49 & 0.81 & 0.99 & 1.00 & 1.00 & 1.00 & 1.00 & 1.00 \\
 & 43 & 290 & 7.78 & 29.44 & 34.44 & 35.00 & 35.00 & 35.00 & 35.00 & 35.00 & 0.49 & 0.79 & 0.93 & 0.93 & 0.93 & 0.93 & 0.93 & 0.93 \\
 & 44 & 240 & 7.78 & 26.11 & 33.33 & 33.89 & 33.89 & 33.89 & 33.89 & 33.89 & 0.49 & 0.83 & 1.05 & 1.08 & 1.08 & 1.08 & 1.08 & 1.08 \\
\addlinespace[2pt]
Nemotron 2k & 42 & 240 & 2.78 & 17.78 & 36.67 & 46.11 & 48.89 & 48.89 & 48.89 & 48.89 & 0.51 & 0.95 & 1.58 & 2.04 & 2.16 & 2.16 & 2.16 & 2.16 \\
 & 43 & 290 & 2.78 & 17.22 & 36.11 & 45.56 & 46.67 & 46.67 & 46.67 & 46.67 & 0.51 & 0.94 & 1.46 & 1.68 & 1.71 & 1.75 & 1.84 & 2.02 \\
 & 44 & 280 & 1.11 & 12.78 & 35.00 & 43.33 & 43.33 & 43.33 & 43.33 & 43.33 & 0.51 & 0.96 & 1.54 & 1.89 & 1.94 & 1.99 & 2.08 & 2.26 \\
\bottomrule
\end{tabular}
\end{table}

\begin{table}[!ht]
\centering
\scriptsize
\setlength{\tabcolsep}{1.8pt}
\renewcommand{\arraystretch}{0.9}
\caption{Final step-300 checkpoints for every training seed. Accuracy and mean response length at each evaluation budget use the same definitions and units as Table~\ref{tab:multiseed-selected}.}
\label{tab:multiseed-final}
\begin{tabular}{@{}lrr*{8}{r}@{\hspace{8pt}}*{8}{r}@{}}
\toprule
Condition & Seed & Step & \multicolumn{8}{c}{Accuracy (\%)} & \multicolumn{8}{c}{Mean response length (k tokens)} \\
\cmidrule(lr){4-11}\cmidrule(lr){12-19}
& & & 512 & 1k & 2k & 4k & 8k & 16k & 32k & 64k & 512 & 1k & 2k & 4k & 8k & 16k & 32k & 64k \\
\midrule
Qwen 4k & 42 & 300 & 0.00 & 0.00 & 16.67 & 34.44 & 43.89 & 47.78 & 47.78 & 47.78 & 0.51 & 1.00 & 1.93 & 3.25 & 4.67 & 5.34 & 5.44 & 5.44 \\
 & 43 & 300 & 0.00 & 0.56 & 22.78 & 41.67 & 47.78 & 50.00 & 50.56 & 50.56 & 0.51 & 1.00 & 1.91 & 3.08 & 4.26 & 4.72 & 4.76 & 4.76 \\
 & 44 & 300 & 0.00 & 3.89 & 21.11 & 37.78 & 46.67 & 51.11 & 51.67 & 51.67 & 0.51 & 0.99 & 1.86 & 3.06 & 4.34 & 4.93 & 4.95 & 4.95 \\
\addlinespace[2pt]
Qwen 8k & 42 & 300 & 0.00 & 0.00 & 2.78 & 27.78 & 55.56 & 55.56 & 55.56 & 55.56 & 0.51 & 1.00 & 1.99 & 3.61 & 4.95 & 5.02 & 5.11 & 5.28 \\
 & 43 & 300 & 0.00 & 0.00 & 0.00 & 18.33 & 50.00 & 53.89 & 53.89 & 53.89 & 0.51 & 1.00 & 2.00 & 3.83 & 6.11 & 6.47 & 6.47 & 6.47 \\
 & 44 & 300 & 0.00 & 0.00 & 3.33 & 25.00 & 51.67 & 54.44 & 54.44 & 54.44 & 0.51 & 1.00 & 1.99 & 3.72 & 5.67 & 5.83 & 5.83 & 5.83 \\
\addlinespace[2pt]
Gemma 1k & 42 & 300 & 2.22 & 8.89 & 11.67 & 12.22 & 12.22 & 12.22 & 12.22 & 12.22 & 0.50 & 0.83 & 1.02 & 1.04 & 1.04 & 1.04 & 1.04 & 1.04 \\
 & 43 & 300 & 3.89 & 13.33 & 17.22 & 17.78 & 17.78 & 17.78 & 17.78 & 17.78 & 0.51 & 0.87 & 1.05 & 1.06 & 1.06 & 1.06 & 1.06 & 1.06 \\
 & 44 & 300 & 2.22 & 15.56 & 24.44 & 24.44 & 24.44 & 24.44 & 24.44 & 24.44 & 0.51 & 0.89 & 1.09 & 1.11 & 1.11 & 1.11 & 1.11 & 1.11 \\
\addlinespace[2pt]
Gemma 2k & 42 & 300 & 0.00 & 3.33 & 17.22 & 20.00 & 20.00 & 20.00 & 20.00 & 20.00 & 0.51 & 0.99 & 1.51 & 1.57 & 1.57 & 1.57 & 1.57 & 1.57 \\
 & 43 & 300 & 0.00 & 2.78 & 12.78 & 17.78 & 17.78 & 17.78 & 17.78 & 17.78 & 0.51 & 0.98 & 1.67 & 1.99 & 2.00 & 2.00 & 2.00 & 2.00 \\
 & 44 & 300 & 0.00 & 5.56 & 18.33 & 18.89 & 18.89 & 18.89 & 18.89 & 18.89 & 0.51 & 0.99 & 1.61 & 1.75 & 1.75 & 1.75 & 1.75 & 1.75 \\
\addlinespace[2pt]
Nemotron 1k & 42 & 300 & 7.78 & 27.22 & 33.33 & 34.44 & 34.44 & 34.44 & 34.44 & 34.44 & 0.50 & 0.82 & 1.01 & 1.04 & 1.04 & 1.04 & 1.04 & 1.04 \\
 & 43 & 300 & 8.33 & 26.67 & 34.44 & 34.44 & 34.44 & 34.44 & 34.44 & 34.44 & 0.49 & 0.80 & 0.94 & 0.94 & 0.94 & 0.94 & 0.94 & 0.94 \\
 & 44 & 300 & 7.78 & 24.44 & 30.00 & 30.00 & 30.00 & 30.00 & 30.00 & 30.00 & 0.50 & 0.82 & 0.98 & 0.98 & 0.98 & 0.98 & 0.98 & 0.98 \\
\addlinespace[2pt]
Nemotron 2k & 42 & 300 & 4.44 & 16.67 & 36.11 & 46.11 & 48.33 & 48.33 & 48.33 & 48.33 & 0.51 & 0.95 & 1.58 & 2.05 & 2.15 & 2.15 & 2.15 & 2.15 \\
 & 43 & 300 & 2.78 & 15.00 & 37.22 & 42.22 & 42.22 & 42.22 & 42.22 & 42.22 & 0.51 & 0.94 & 1.45 & 1.68 & 1.73 & 1.82 & 2.00 & 2.35 \\
 & 44 & 300 & 1.11 & 11.11 & 35.56 & 44.44 & 45.00 & 45.00 & 45.00 & 45.00 & 0.51 & 0.97 & 1.53 & 1.84 & 1.86 & 1.86 & 1.86 & 1.86 \\
\bottomrule
\end{tabular}
\end{table}

For Qwen 4k seed 44, Gemma 1k seed 43, and Gemma 2k seed 42, the selected checkpoint is also the final checkpoint; the two tables therefore contain 33 distinct trained checkpoints. The accompanying data files retain per-seed accuracy, length, and truncation curves at every budget.